\PassOptionsToPackage{nonamebreak}{natbib}
\documentclass{article} 
\usepackage{iclr2027_conference,times}

\usepackage{algorithm}
\usepackage{algpseudocode}
\usepackage{amsmath,amssymb}
\usepackage{booktabs}
\usepackage{multirow}
\usepackage{adjustbox}
\usepackage[table]{xcolor}
\usepackage[most]{tcolorbox}
\usepackage{xspace}
\usepackage{enumitem}

\usepackage{amsmath,amsfonts,bm}

\def\eqref#1{equation~\ref{#1}}

\def\1{\bm{1}}

\DeclareMathAlphabet{\mathsfit}{\encodingdefault}{\sfdefault}{m}{sl}
\SetMathAlphabet{\mathsfit}{bold}{\encodingdefault}{\sfdefault}{bx}{n}

\usepackage{hyperref}
\usepackage{url}

\title{MagicBeanStalk: Escaping the RLVR Zero Advantage Trap 
with Recursive Data Synthesis}
 
\definecolor{bestcell}{HTML}{E8F5E9}    
\definecolor{hdrcolor}{HTML}{F5F5F5}    

\iclrfinalcopy
\author{
Ximing Lu$^{1}$\thanks{Core contributors.},
Aviral Kumar$^{2,4}$, Jan Kautz$^{4}$,
Yejin Choi$^{3}$, David Acuna$^{4,*}$ \\
$^{1}$University of Washington \quad
$^{2}$Carnegie Mellon University \\
$^{3}$Stanford University \quad $^{4}$NVIDIA \\
\texttt{lux32@cs.washington.edu},
\texttt{dacunamarrer@nvidia.com}
}

\begin{document}
\raggedbottom
\newcommand{\gptoss}{\texttt{gpt-oss-120b}\xspace}

\maketitle
\begin{abstract}
Data-centric scaling of RLVR has largely focused on collecting or generating verifiable training problems.
However, corpus size alone does not determine whether the dataset provides a useful learning signal for a target policy. To guide the synthesis of effective RLVR data,  we analyze the  RLVR objective and experimentally characterize dataset effectiveness as a function of its problems, the target policy, and the training rollout budget. 
Under this characterization, we show that a large open-source math corpus pooled from widely used RLVR datasets falls almost entirely into a \emph{zero-advantage trap}: for a fixed target policy and rollout budget, roughly \textbf{88\%} of its problems are solved in every rollout or failed in every rollout.
Filtering concentrates what remains, but it can only select, discarding the hard problems. 
Effective synthetic problems should therefore require the reasoning operations and  compositions required to solve those hard problems.
Yet their difficulty must be calibrated, so that some target-policy rollouts succeed, some fail, and at least one complete correct solution fits within the training budget.
Prompting a teacher to vary hard problems directly does not achieve this. 
Guided by these empirical design principles, we introduce \textbf{MagicBeanStalk}, a top-down 
decomposition and bottom-up recursive data-synthesis algorithm that grows
seed problems into families of substantively new, policy- and
budget-calibrated training problems while preserving the seeds' reasoning
blueprints. 
Across 8 math reasoning benchmarks, continual RLVR
on MagicBeanStalk problems improves gpt-oss-120b from 61.2 to \textbf{69.8} (\textbf{+8.6}) and DeepSeek~v4~Flash from 72.4 to \textbf{77.9} (\textbf{+5.4}) on a 16K-token budget. Training on the full 78K source corpus improves gpt-oss-120b by only {+1.1} and its best filtered subset by +5.8.
In finance, 
we  synthesize over 14K verifiable training problems from only 340 solutions and  improve
Qwen3 by \textbf{+6.9} and \textbf{+2.4} points 
on SecQue and FinanceBench, respectively.

\end{abstract}
\providecommand{\teachers}{\mathcal{T}}                         
\providecommand{\model}{\mathcal{\pi}}                            

\providecommand{\problem}{p}                                    
\providecommand{\problemspace}{\mathcal{P}}
\providecommand{\question}{q}                                   
\providecommand{\questionspace}{\mathcal{Q}}
\providecommand{\answerspace}{\mathcal{A}}
\providecommand{\doc}{d}                                        
\providecommand{\docspace}{\mathcal{D}}

\providecommand{\trace}{\tau}
\providecommand{\tracespace}{\mathcal{R}}                       
\providecommand{\extract}{\mathrm{extract}}
\providecommand{\verified}{\mathrm{verified}}

\providecommand{\tree}{\mathcal{G}}                             
\providecommand{\childset}{\mathcal{C}}

\providecommand{\acc}{\mathrm{acc}}
\providecommand{\condacc}{\mathrm{acc}^{\mathrm{cond}}}
\providecommand{\goldilocks}{\mathcal{Q}_{\mathrm{gold}}}

\begin{figure}[t]
    \centering
    \includegraphics[width=0.99\textwidth,
    keepaspectratio,
          trim={0 0 0 0},
          clip]
          {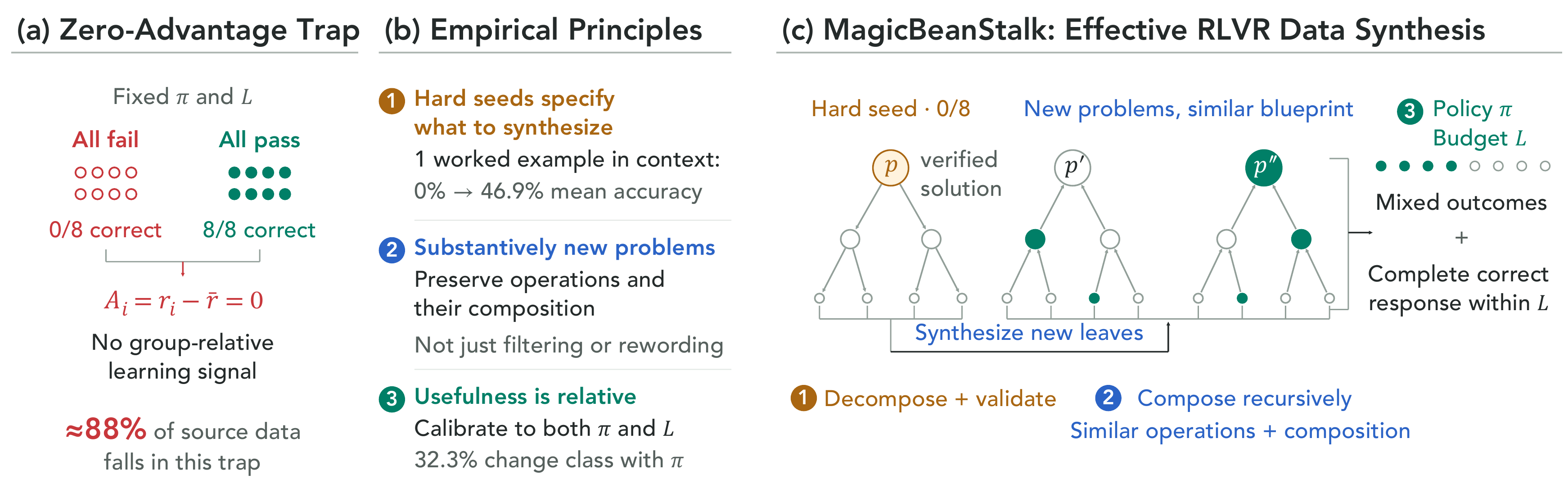}
 
\caption{\textbf{MagicBeanStalk synthesizes RLVR data from hard seeds to escape the zero-advantage trap.}
\textbf{(a)} For a fixed policy $\pi$ and budget $L$, all-failure and all-success rollout groups yield zero advantage; $\approx 88\%$ of a $78$K math corpus falls in this trap for \gptoss at 32K.
\textbf{(b)} Three empirical principles: hard seeds specify what to synthesize; synthetic problems must be substantively new yet preserve the seed's reasoning operations and their composition; and usefulness is relative to $\pi$ and $L$.
\textbf{(c)} MBS applies them in order: the seed's verified teacher trace is decomposed top-down into subproblems that help a student solve their parent (1); harder new leaves are synthesized and recursively composed bottom-up into parallel trees $p'$, $p''$ (2); synthetic root problems with mixed target-policy rollouts and a complete correct response within $L$ are kept (3).
\vspace{-4mm}
}

    \label{fig:magicbeanstalk_overview}
\end{figure}

\section{Introduction}

Reinforcement learning with verifiable rewards (RLVR) has become key for improving reasoning models, but as policies become stronger, scaling
RLVR increasingly becomes a data  problem  requiring a continual supply of
verifiable data. Existing efforts have expanded this supply by
collecting human-authored problems for math, coding and other domains~\citep{chen2025acereason,deepscaler2025,
albalak2025bigmathlargescalehighqualitymath,
cui2025process,
lu2025scp116khighqualityproblemsolutiondataset}
, as well as by developing procedural
environments that can generate multiple problems per environment~\citep{
stojanovski2025reasoninggymreasoningenvironments,lacombe2025reasoning,
Zeng2025RLVESU}
. However, sourcing high-quality human problems
become increasingly hard
as we run out of internet data. 
And  procedural environments are typically hand-engineered, 
limited to a set of domains, and often
produce problems that vary surface details 
within a fixed
reasoning structure. As a result, recent approaches synthesize surrogate tasks from Internet data~\citep{lu2026golden} , yet more 
data
does not necessarily ensure  the problems
provide useful learning signal for a target policy and training budget.

In this work, we return to the binary-reward RLVR objective and characterize
dataset effectiveness as a function of three interacting components: the
problems it contains, the target policy, and the training rollout budget. 
Using this characterization and a set of 
controlled experiments, we design key principles to guide the synthesis of effective RLVR data. We demonstrate that more data does not necessarily lead to better downstream performance, and a large corpus can provide
little learning signal when most problems fall into what we call the
\textbf{zero-advantage trap} (i.e. problems are either too easy or too hard for the
target policy under the training budget). Specifically, in a corpus of approximately
$78K$ open-source math problems, roughly $88\%$ fall into this trap for \gptoss at a 32K rollout budget. Filtering alone is limited to selection, and disregards the hard problems that sits at the frontier of the target policy and 
expose  the reasoning capabilities the policy lacks.

Our key insight is to repurpose the hard
problems  as seeds from which to grow substantively new problems whose solutions
require the same reasoning operations, but whose difficulty is calibrated to
produce mixed rollout outcomes and admit at least one complete correct solution
within the training budget. 
However, the obvious baseline---directly prompting
a strong teacher to produce variations of the hard seed---
performs no better than training on the best problems already present in the corpus.

This motivates a more structured construction, \textbf{MagicBeanStalk} (MBS). MBS is  a top-down
decomposition and bottom-up recursive synthesis method that grows a small collection of seeds into families of policy- and budget-calibrated RLVR problems.
MBS decomposes each seed into a validated tree of simpler subproblems, synthesizes substantively new leaf problems, and recombines them
while preserving the seed's reasoning blueprint~(Figure~\ref{fig:magicbeanstalk_overview}). 
MBS escapes the trap where direct synthesis does not: from hard seeds
on which \gptoss scores $0/8$, it produces in-budget Goldilocks problems for
$\mathbf{74.7\%}$ of seeds versus $43.8\%$ for direct synthesis.
In a matched training control on a shared seed population, MBS reaches 69.2, versus 63.9 for direct synthesis and 65.3 for training on the seeds themselves.
On \gptoss, continual RLVR raises the avg across 8 math reasoning
benchmarks (AIME~24/25, AMC, HMMT-F/N, Minerva, Olympiad, Omega-500) from $61.2$
to $\mathbf{69.8}$ ($\mathbf{+8.6}$ points) at a 16K-token budget. The same
data applied to DeepSeek~v4~Flash raises its average from $72.4$ to
$\mathbf{77.9}$ ($\mathbf{+5.4}$ points) under the same budget.
Further, 
MBS gains persist at 24K and 32K and   on MathArenaApex~\citep{matharena_apex} they exceed the best baseline's gain at every budget.
Beyond math, we also apply MBS to finance, a specialized domain where annotation is expensive and verifiable RL data is scarce. Starting from
only $340$ expert-curated seed questions, MBS synthesizes over
$14{,}000$ grounded financial reasoning problems. Continual RLVR on this
data improves Qwen3-4B by $\mathbf{+6.9}$ pass@1 on SecQue and
$\mathbf{+2.4}$ on FinanceBench.

\begin{figure*}[t]
  \centering

  \begin{minipage}[t]{0.49\textwidth}
      \centering
      \includegraphics[
          width=\linewidth,
          height=4.3cm,
          keepaspectratio,
          trim={0 0 0 0},
          clip
      ]{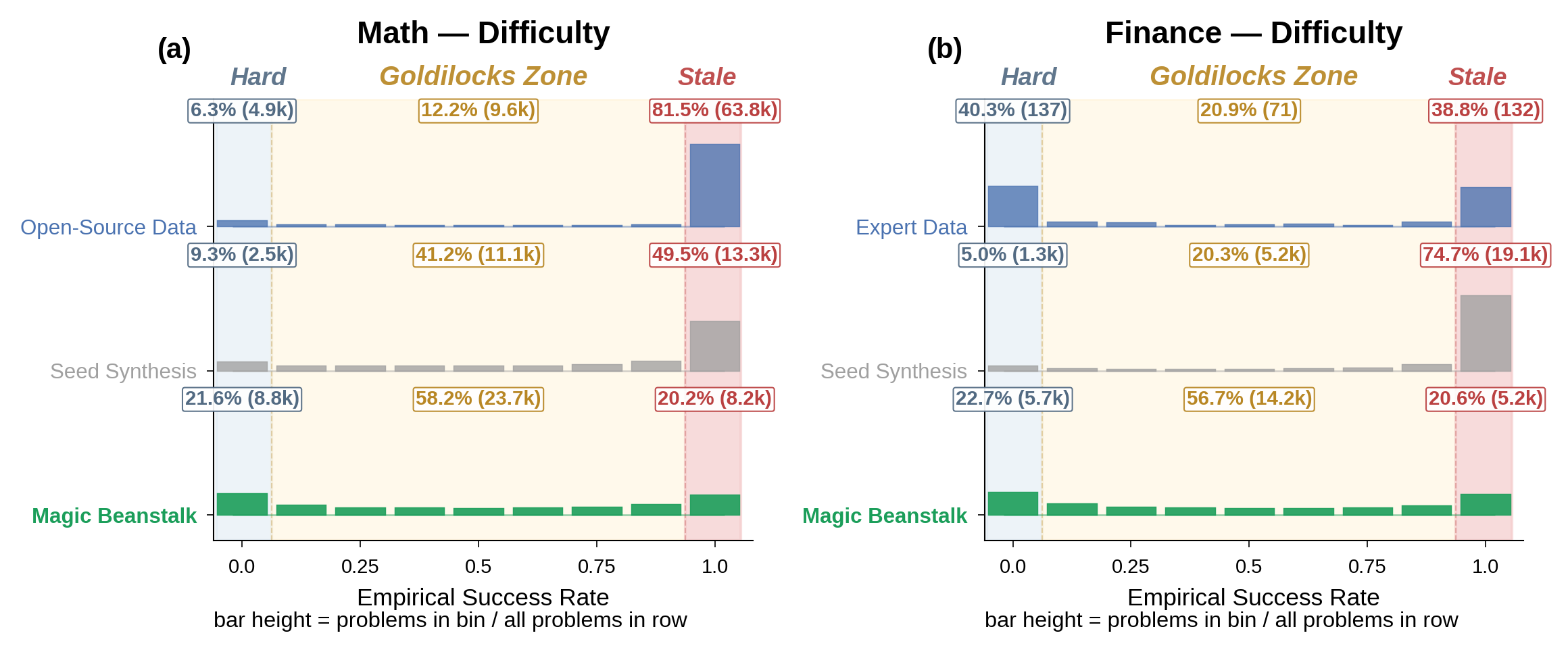}

  \end{minipage}
  \hfill
  \begin{minipage}[t]{0.5\textwidth}
      \centering
      \includegraphics[
          width=\linewidth,
          height=4.8cm,
          keepaspectratio,
          trim={30 5 5 0},
          clip
      ]{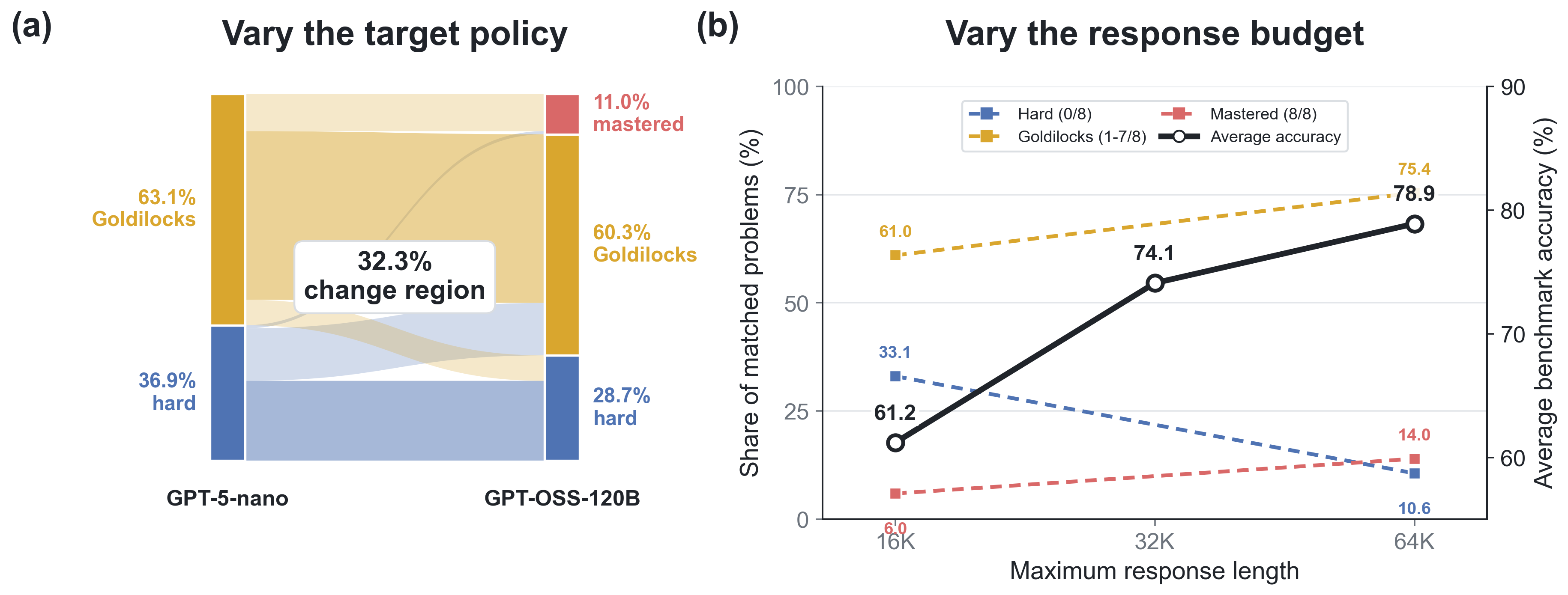}

  \end{minipage}

  \caption{
      \textbf{Effective RLVR data synthesis must account for the problem, target policy, and response budget.}
      \textbf{Difficulty distributions (left):} Empirical success-rate
      distributions from 8 \gptoss rollouts per problem in math
      and finance. We denote problems with $0/8$, $1$--$7/8$, and $8/8$
      correct responses as \emph{hard}, \emph{Goldilocks}, and \emph{stale/mastered},
      respectively; each bar height is the fraction of problems in that row.
       Roughly \textbf{$\mathbf{88\%}$ of the corpus falls in the zero-advantage trap} providing no learning signal,
      whereas
      MagicBeanStalk places $58.2\%$ of its math problems and $56.7\%$
      of its finance problems in the Goldilocks band.
      \textbf{Policy and budget dependence (right):} The two panels on (b) use the
 problem subset that GPT-5-nano does not solve in every rollout  excluding most of the stale mass on the left. Holding the problems and
32K response limit fixed, changing the target policy reclassifies $32.3\%$ of
problems. Holding \gptoss fixed, increasing the
      response limit from 16K to 64K reduces the fraction of hard problems from $33.1\%$ to $10.6\%$. Black curve reports avg accuracy.
      Difficulty depends on the problem, policy, and
      rollout budget.
  }
  \label{fig:difficulty_controls_main}
  \vspace{-4mm}
\end{figure*}

\vspace{-2mm}
\section{Design Principles for Effective RLVR Data Synthesis}
\label{sec:useful_rlvr_data}

To derive design principles for useful RLVR data synthesis, we first
characterize when a problem provides learning signal for a target policy under
a fixed training rollout budget. Consider a verified problem $(\problem,a^*)$, a target policy $\model$, and a
rollout budget $L$. We sample $N$ responses
$(\trace_i,a_i) \sim \model(\cdot\mid \problem;L)$, where $\trace_i$ is the
reasoning trace and $a_i$ is the final answer. We score each answer with a
binary verifier, $r_i=R(\problem,a_i)\in\{0,1\}$, using the reference answer
$a^*$. The empirical success rate and the corresponding
group-relative advantage estimate are:
\begin{equation}
\vspace{-3mm}
\bar r_{\pi,L}(\problem)
= \frac{1}{N}\sum_{j=1}^{N}r_j,
\qquad
\hat A_i(\problem)
= r_i(p)-\bar r_{\pi,L}(\problem).
\label{eq:advatange}
\vspace{-3mm}
\end{equation}

\begin{figure*}[t]
  \centering

  \begin{minipage}[t]{0.49\textwidth}
      \centering
      \includegraphics[
          width=\linewidth,
          height=4.8cm,
          keepaspectratio
      ]{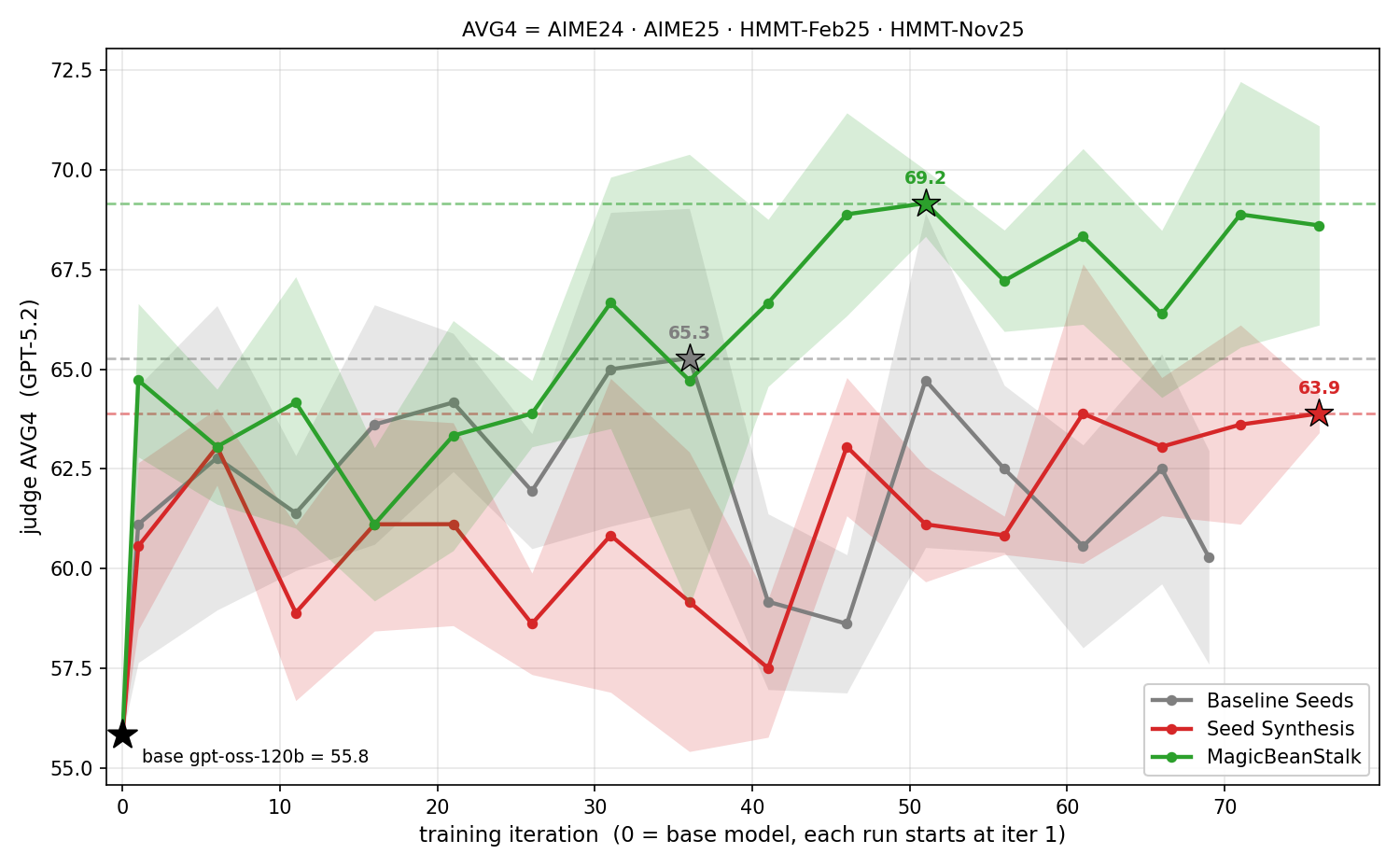}

      \vspace{0.1mm}
      \small\textbf{(a)} Downstream training dynamics
  \end{minipage}
  \hfill
  \begin{minipage}[t]{0.49\textwidth}
      \centering
      \includegraphics[
          width=\linewidth,
          height=4.8cm,
          keepaspectratio
      ]{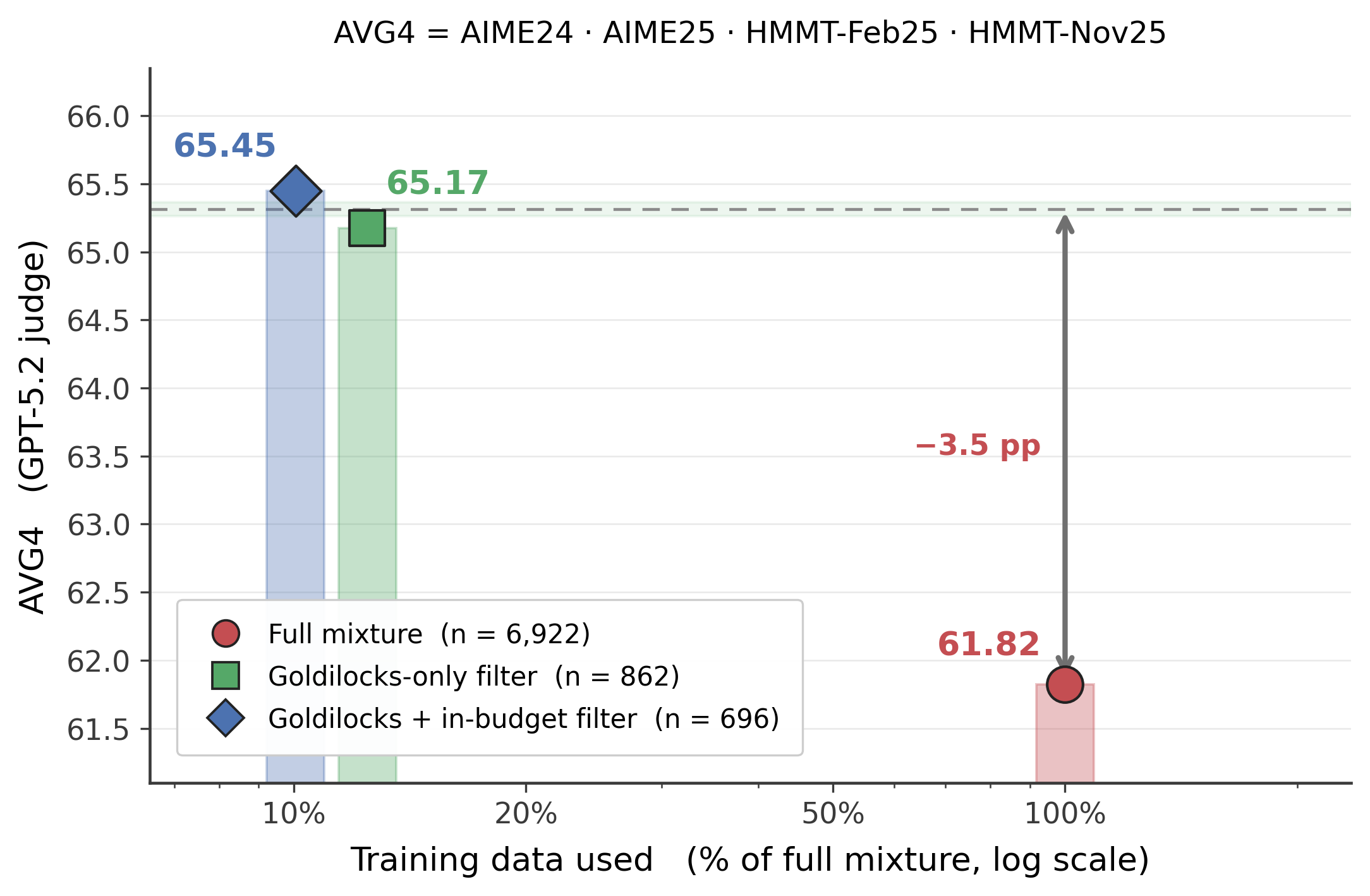}

      \vspace{0.1mm}
      \small\textbf{(b)} A large corpus can provide little signal

  \end{minipage}

 \caption{
      \textbf{MBS improves beyond seed training and direct synthesis. Filtering shows that more data alone is not enough but discards the hard problems the target policy is not able to solve yet.}
      \textbf{(a)} Starting from the same
      base policy, we compare one epoch of training on fixed
      seed problems and the same amount of qualified seed-synthesis 
      and MBS problems. All derived from the same fixed seed
      population.
      Seed synthesis does not surpass
      seed-only training, whereas MBS provides a clear additional
      gain. In this control experiment, the seed population mixes goldilocks and hard problems, so that seed-only training is a meaningful baseline.
      \textbf{(b)} Holding the base
      policy, training recipe, and evaluation fixed, we train on a
      $6{,}922$-problem proxy that preserves the source mixture's difficulty
      distribution, on its $862$-problem Goldilocks subset, and on the
      $696$-problem subset that additionally demonstrates in-budget
      solvability. Both filtered subsets outperform the full proxy while using
      fewer problems.
  }
  \vspace{-1.3mm}
  \label{fig:seed_synthesis_training_main}
\end{figure*}

In a mixed group, $\hat A_i$ is positive for a successful response and
negative for an unsuccessful response. When all responses receive the same
reward, $\hat A_i=0$ for every response, and the sampled group provides no
within-problem advantage signal. 
 We refer to this failure mode as the \textbf{zero-advantage trap}. These are empirical finite-sample classifications depending on rollout counts, they do not establish absolute inability/mastery.
Further, we call
problems satisfying $0<\bar r_{\pi,L}(\problem)<1$ the policy's
\emph{Goldilocks band}. This implies some rollouts succeed and others fail, yielding a
nonzero group-relative learning signal. The definition is explicitly
relative to the target policy $\pi$ and the rollout budget $L$ and counts. 
\textbf{The zero advantage trap and the goldilock bands. More data does not always means better performance.}
We characterize the learning signal
available in a large corpus of open-source mathematical reasoning problems using the success rate.
The corpus combines {problem from
  DAPO-Math-17K~\citep{Yu2025DAPO},
  Skywork-OR1-RL-Data~\citep{he2025skywork}, Turing
  Open-RL~\citep{turing_2026_open_rl}, and a subset of StackMathQA~\cite{zhang2024stackmathqa},
} and contains
approximately $78{,}000$ problems. For each problem, we sample $N=8$ responses
from \gptoss using rollout budget $L$.
Despite the size of the corpus, only $12.2\%$ of its problems fall within the
observed Goldilocks band ($L=32K$). As shown in Figure~\ref{fig:difficulty_controls_main},
most problems are solved in all rollouts and therefore have
empirical success rate $\bar r_{\pi,L}(\problem)=1$ while a $6.3\%$ are not solved in any rollout. 
Therefore, roughly \textbf{$\mathbf{88\%}$ of the corpus falls in the zero-advantage trap}. They contribute  no group-relative advantage.
A matched downstream control experiment
further
shows that the Goldilocks subset and a stricter subset that additionally
demonstrates in-budget solvability outperform a proxy preserving the full
corpus distribution, despite using an order of magnitude fewer problems.
(Figure~\ref{fig:seed_synthesis_training_main}(b))
Thus, more RLVR data does not necessarily means better downstream performance.  (Appendix~\ref{app:abundant_little_signal} for more)

\textbf{Difficulty is relative to the policy and training budget}.
Crucially, whether a problem provides useful signal depends on both the target
policy and the rollout budget. Holding the problems and a 32K response limit
fixed, changing the target policy reclassifies $32.3\%$ of problems among the
zero-success, Goldilocks, and fully solved regions. Holding the policy fixed,
increasing the response limit from 16K to 64K reduces the fraction of
apparently hard problems from $33.1\%$ to $10.6\%$
(Figure~\ref{fig:difficulty_controls_main}). A truncated response therefore does
not establish that the policy lacks the required reasoning capability. It
establishes only that no complete solution was produced within budget. 
We say
that a problem has demonstrated \emph{in-budget solvability} if at least one
correct response terminates normally (i.e. without reaching the rollout limit).
Difficulty is consequently a property of the tuple.
Therefore, generated problems should be
calibrated to the target policy and training rollout budget, producing both
successful and unsuccessful rollouts while admitting at least one complete
correct response within budget. Filtering alone, however, can only select
from problems a corpus already contains. Filtering cannot create new problems targeting capabilities absent from the filtered subset.

\textbf{Hard problems are specifications for what data to synthesize.}
\label{sec:hard_seeds_frontier}
Hard problems play a paradoxical role in RLVR. 
Problems with zero observed success provide no group-relative advantage in the sampled group, yet their verified solutions identify reasoning operations and compositions to target through synthesis.
We refer to the essential reasoning operations exposed by a verified
solution and how those operations are composed as
\emph{reasoning blueprints}. 
We define a
\emph{verified hard problem} as a solvable problem for which
$\bar r_{\model,L}(\problem)=0$ and the sampled responses terminate normally.
%
The termination requirement ensures that the observed failures 
are not artifact of the rollout budget.
To test whether these blueprints provide actionable specifications, we study $98$ hard math seeds and obtain verified traces from a pool of 5 frontier models.
For each seed, we generate one simpler subproblem instantiating some of its blueprint operations and provide its verified solution as a single in-context example. We verify that it does not reveal the original solution. This intervention enables the target policy to solve at least one rollout for $78/98$ seeds ($79.6\%$) and raises mean accuracy on the original problems from $0\%$ to $46.9\%$.  Thus, hard problems are not merely training examples. Their verified solutions provide specifications of what skill a synthetic problem   should target. Moreover, among the $43$ seeds with at least two verified blueprints obtained with different teachers, $90.7\%$ produce the same blueprint or a compatible variant. This further suggests that the blueprint primarily reflects the problem's structure rather than a particular teacher style. (See Appendix~\ref{app:hard_seeds_frontier} for more)

\begin{figure*}[t]
  \centering

  \begin{minipage}[t]{0.39\textwidth}
      \centering
      \includegraphics[
          width=\linewidth,
          keepaspectratio,
          trim={0 0 0 0},
          clip
      ]{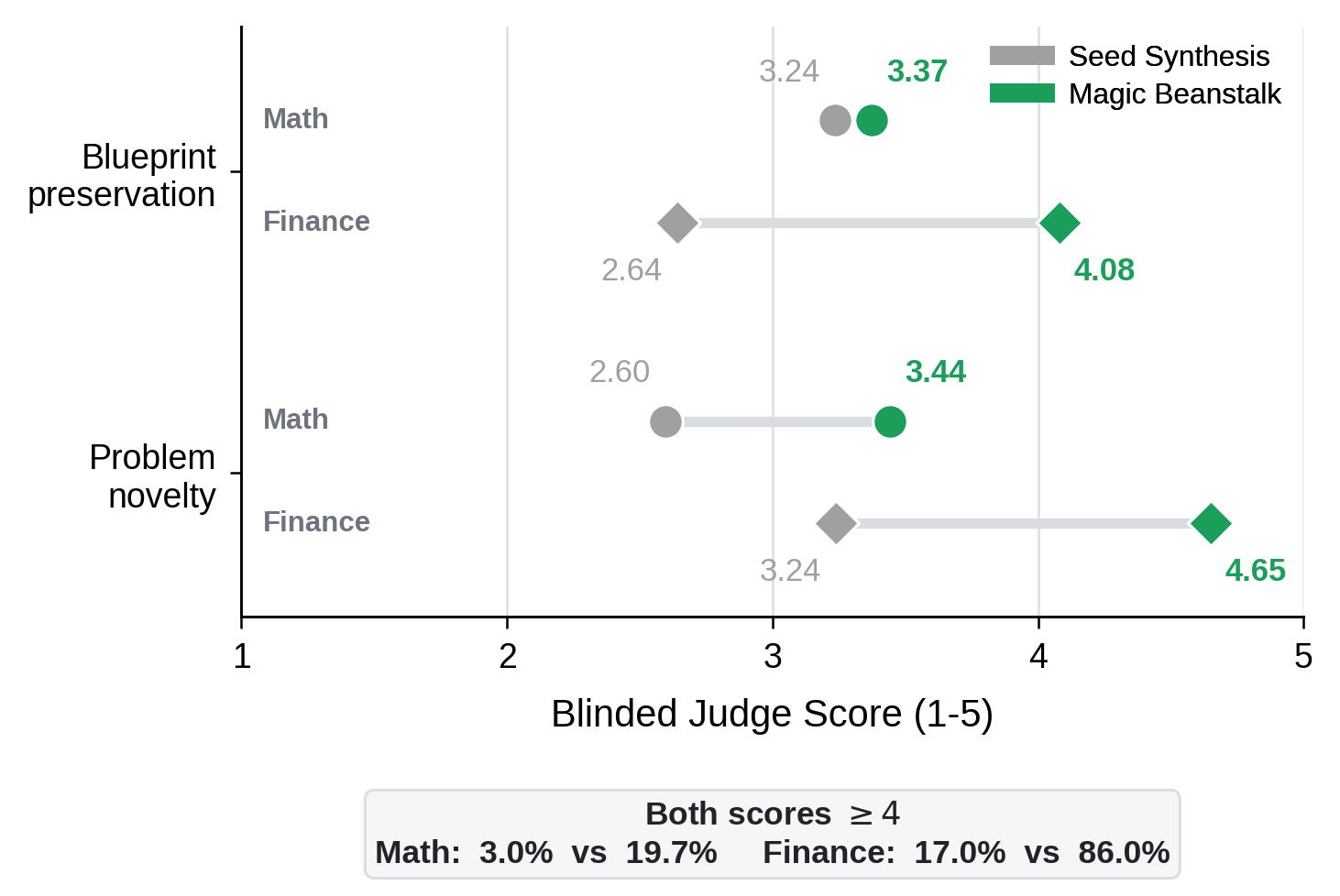}

      \vspace{1mm}
      \small\textbf{(a)} Blueprint preservation and novelty
  \end{minipage}
  \hfill
  \begin{minipage}[t]{0.59\textwidth}
      \centering
      \includegraphics[
          width=\linewidth,
          keepaspectratio,
          trim={0 0 0 0},
          clip
      ]{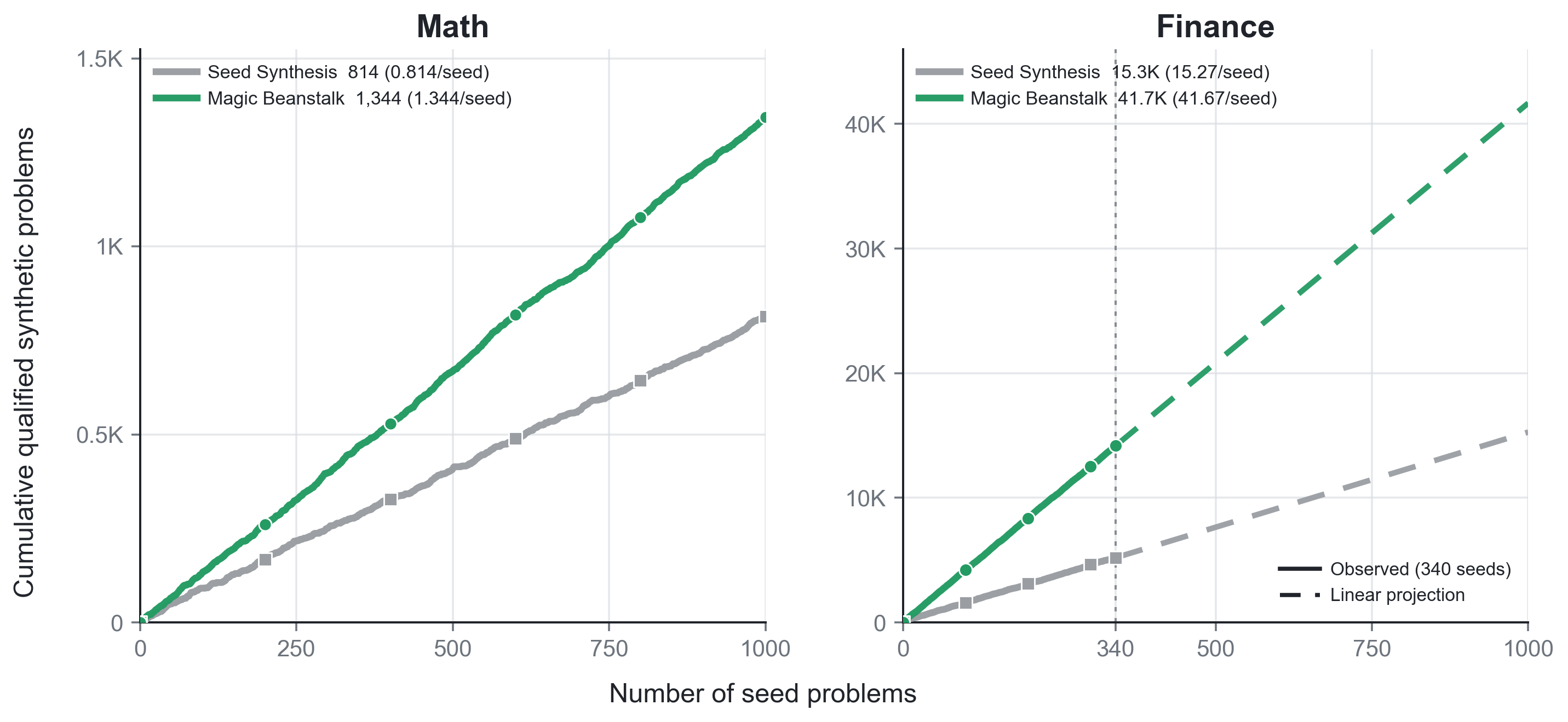}

      \small\textbf{(b)} Qualified problems per source seed
  \end{minipage}

  \caption{
      \textbf{Direct prompting a teacher to generate variants is limited by both the problems it
      constructs and the number of useful candidates it yields.}
      \textbf{Left:} We evaluate $300$ same-seed pairs in
      math and $100$ in finance for each approach. A judge scores each generated problem
      for reasoning-blueprint preservation and problem novelty.
      Blueprint preservation measures if the solutions to the synthetic problem requires
      the seed's essential reasoning operations and dependencies. Novelty
      measures if the synthetic problem is new rather than a
      paraphrase, contextual rewrite, or numerical substitution. Higher is
      better.
      \textbf{Right:} Cumulative qualified problems as source seeds are added,
      (both methods evaluated on the same seeds).
      Qualification requires $1$--$7/8$ correct 
      and at least one correct response that terminates normally
      . In
      finance,  we use same qualification. 
      Finance curves are observed
      through 340 seeds and linearly projected thereafter. MBS
      produces more qualified descendants per seed while more often
      preserving the blueprint in a  new problem.
  }
  \label{fig:seed_synthesis_analysis}
  \vspace{-4mm}
\end{figure*}

\textbf{Direct prompting a teacher to generate variations is not enough.}
Filtering can only select from problems a corpus already contains
while hard seeds specify the
reasoning the policy is missing 
and left out during filtering. 
A simple way to synthesize problem starting with hard seeds is
to provide the problem, reasoning trajectory, and its verified solution to a strong teacher
. Then perform rejection sampling on quality control, goldilocks and in-budget solvability. 
This baseline is straightforward and we experimentally observed it often retains the mathematical techniques used by the seed. In a controlled set of experiments, however, we observed \textbf{seed synthesis provides no clear gain beyond the seeds.} 
Figure~\ref{fig:seed_synthesis_training_main}(a) shows the training curves. 
We experimentally found two main reasons for this. 
\textbf{(1) seed synthesis preserves techniques but often makes shallow
variants.}
We study this behavior with a blinded judge study over $300$ generated
problems in math and $100$ in finance.
Figure~\ref{fig:seed_synthesis_analysis} (left) shows that
seed synthesis often retains recognizable reasoning techniques, particularly
in math, but frequently does so through a shallow variation of the
original problem rather than a substantively new instantiation. And
\textbf{(2) post-hoc filtering calibrates difficulty but cannot repair the
candidate pool.}
Filtering is subtractive
and cannot turn a shallow variant into a
substantively new problem, nor produce a candidate for a seed that yielded
none. Seed synthesis consequently returns below one qualified problem per seed
(Figure~\ref{fig:seed_synthesis_analysis}, right). We study the full yield and
seed-coverage comparison in Section~\ref{sec:exp_data_quality}.
Also, see appendix~\ref{app:direct_seed_synthesis}
 for more details.

\textbf{Takeaway.} Three principles follow. First, hard problems are specifications. Their verified solutions name the reasoning operations and skills synthesis should target. Second, synthesis must do what filtering cannot and should be complementary to dynamic sampling. Synthetic problems should be substantively new and preserve the hard seed's reasoning operations and their composition. Third, usefulness is relative. A problem carries signal  for a given target policy and rollout budget, so generated problems must be calibrated to both. They should yield mixed target-policy rollouts with a complete correct response within the training budget. 
%

 %
\section{MagicBeanStalk: Effective Data Synthesis from Hard Seeds}
\label{main:MBS}
\vspace{-2mm}
We now introduce MagicBeanStalk (MBS), a data synthesis algorithm motivated by the findings in  in~\ref{sec:useful_rlvr_data}. Each stage in the following subsections is inspired by those principles. 

\subsection{Preliminaries and Notation}

We assume access to a collection of \textbf{teacher} models $\teachers := \{T_1, T_2, \ldots, T_M\}$. In our setting, these are strong reasoning LLMs used to
solve, decompose, synthesize, and verify problems. 
Additionally, we define an ordered pool of \textbf{student} models
$\mathcal{S}:=\{S_1,S_2,\ldots,S_J\}$, ranked from strongest ($S_1$) to weakest ($S_J$), and a  target   policy $\model$. In our setting, $\model$ is part of the student pool. 
A problem is a tuple $(\problem, a^*)$  where $\problem$ is a question and $a^* \in \answerspace$ is its ground-truth answer. In document-grounded
settings (e.g.~finance), we  let the $\problem=(\question,\doc) $ where  $\doc$ is a document
snippet that contains the information required to answer $\question$.
\textbf{Decomposition tree.} 
A \textbf{decomposition tree}
$\tree_\problem := \langle \problem, \trace, a^*, \childset \rangle$ is
rooted at $(\problem, a^*)$ with its verified teacher trace $\trace$, and
recursively contains a set of $K$ \textbf{children}
$\childset = \{\tree_{\problem_1}, \ldots, \tree_{\problem_K}\}$, each itself a
 tree on an easier subproblem. A node $n$ with
$\childset_n = \emptyset$ is a \textbf{leaf}.

\subsection{Rejection-Sample Teacher Traces}

Given a set of verified problems $(\problem, a^*)$, we use a collection of teachers to sample
reasoning traces. Formally, for each problem $\problem$, we iterate through teachers $T \in \teachers$ in order
and, from each, draw up to $N_t$ reasoning traces and answers
$(\trace^{(i)},a) \sim T(\problem)$ with $i \in \{1, \ldots, N_t\}$. A trace
is \textbf{verified} when its final answer matches the GT.
We accept the first trace satisfying $\verified(\trace, a^*) = 1$ and store it as the root's verified trace in
$\tree_\problem$. Here, $\verified$ represents a verification function either answer matching or an LLM-as-a-judge. 
Problems for which no trace verifies after $|\teachers| \cdot N_t$ attempts are discarded. A collection of teacher reduces the number of discarded problems due to solvability. Section~\ref{app:hard_seeds_frontier} motivates why multiples teacher are beneficial for solution coverage.  

%

\subsection{Top-Down Problem Decomposition}
%

%

We treat a verified $(\problem, \trace, a^*)$ as the root of a decomposition tree
$\tree_\problem$. 
The teacher is then asked to identify the key skills and knowledge required to solve $\problem$, and emit $K$ easier self-contained subproblems that collectively teach those skills, each with its own answer and reasoning trace:
  $\{(\problem_k,\trace_k, a_k^* )\}_{k=1}^{K}
  \;\sim\; T(\problem, \trace, a^*).$
Each subproblem is itself rejection-sampled and verified as in the
previous stage. Each verified subproblem becomes a child node $\tree_{\problem_k}$ that is in turn used as a parent in the recursive decomposition. 
In our experiments $K \leq 3$.

\textbf{Does the subproblem help a student solve the harder problem?}
At each node, we evaluate the student pool from weakest to strongest and
select the strongest evaluated student with zero empirical accuracy as the evaluated student.
We then run an \emph{effectivity test}.  Specifically, each evaluated student is given the
$K$ subproblems and their traces as in-context worked examples and re-tested
on the parent. Let $\mathcal{S}_n \subseteq \mathcal{S}$ denote the students
evaluated at node $n$. A decomposition is \textbf{effective} when conditioning raises accuracy for at least one student:
%
%
  $
   \exists S \in \mathcal{S}_n:\quad
 \condacc(S, \problem \mid \{\problem_k, \trace_k, a^*_k \}_k)
 \;>\; \acc(S, \problem).$
%
Here $\acc(S,\problem)$ is the fraction of $N$ sampled solutions whose answer
matches $a^*$.
We keep effective decompositions and recurse on each child. If the
student pool does not show an improvement, we re-prompt the teacher for a fresh decomposition
with the previous failed attempts as negative context, up to $N_d$
retries. Nodes whose decompositions remain ineffective are dropped.
In our experiments $N = 8$ and $N_d = 3$, and we recurse up to
depth $D = 3$. The effectivity test is motivated by the second design principle (See \ref{sec:useful_rlvr_data} and ~\ref{app:hard_seeds_frontier}).

\subsection{Bottom-up synthesis of calibrated problems}
%

%

At this stage, every retained seed has a decomposition tree $\tree_\problem$ rooted at $(\problem,\trace,a^*)$, with verified subproblems at its descendant nodes. 
We now walk this tree bottom-up to synthesize a parallel \emph{synthetic}
tree of new problems that share the original's reasoning structure. 
At each leaf, we explicitly ask the teacher to generate harder variants of the
leaf problem, and recursively generate a synthetic parent problem by combining the
retained synthetic children.
 Formally, a synthetic parent $(\widetilde\problem,\widetilde\trace,\widetilde a^*)$ is obtained recursively from its $K$ synthetic child problems:
$
    (\widetilde\problem,\widetilde\trace,\widetilde a^*)
    \;\sim\;
    T\bigl(
      \{(\widetilde\problem_k,\widetilde\trace_k,
         \widetilde a_k^*)\}_{k=1}^{K}
      \,;\,
  \mathcal{B}(\tree_n)
    \bigr).
$ 
where $\mathcal{B}(\tree_n)$ denotes the original parent–children configuration at the current node. The original $(\text{parent},\text{children})$ tree is provided
to the teacher as a demonstration pattern that the combination should
mirror in logical structure and difficulty. Each candidate synthetic parent is
itself rejection-sampled and verified, and is discarded when its accuracy on
the weakest student $S_J$ exceeds $\alpha_\text{parent}$. Propagating upward,
every node ends up with synthetic versions, yielding a
complete synthetic tree that parallels the original. In our experiments
$\alpha_\text{parent} = 0.625$ and up to $N_c = 10$ child-combination
trials are attempted per parent.

\textbf{Leaf synthesis.} For each leaf
$(\problem_\ell,\trace_\ell,a_\ell^*)$, the teacher samples $N_\ell$
 harder variants of the question:
$
  \{(\widetilde\problem_\ell^{(s)}, \widetilde\trace_\ell^{(s)}, \widetilde a_\ell^{*(s)})\}_{s=1}^{N_\ell}
  \;\sim\; T(\problem_\ell, \trace_\ell, a_\ell^*).
$
We experimentally noted that explicitly requiring \emph{harder} variants
at the leaves was beneficial. Without this constraint, the teacher tends
to default to easy paraphrases that lose both difficulty and diversity
of the synthetic tree.  Variants whose accuracy on the weakest student $S_J$
exceeds $\alpha_\text{leaf}$ are discarded, and we retry up to $R$
rounds with the remaining teachers. In our experiments,
$R = 3$, and 
$\alpha_\text{leaf} = 0.8$.

\textbf{Grounding on new documents.} In document-grounded scenarios
(e.g.,~finance), each synthetic must remain answerable from a document.
We retrieve top-$K_d$ unseen documents from a corpus by embedding
similarity to the seed's document summary, then ask the teacher to
extract from each a relevant snippet $\doc'$ together with an
\emph{entity of interest} $e'$ .
The synthetic tree is then re-grounded on $\doc'$ with every node
threaded around $e'$. In our experiments $K_d \in \{35, 75\}$ documents per seed.
Finally, we retain only synthetic problems that fall in the Goldilocks band of the target policy and that satisfy in-budget solvability.

\begin{table*}[t]
  \centering
  \vspace{-3mm}
  \caption{\textbf{Main math results at 16K output tokens.}
  Pass@1 accuracy (\%) across AIME~24/25, AMC, HMMT-F/N, Minerva, Olympiad,
  and Omega-500. Reference API models use 4 samples per problem and are included as
  context; \gptoss uses 16 and DeepSeek~v4~Flash uses one. AVG is the mean
  across all 8 benchmarks and $\Delta$ is the within-base-model improvement.}
  \label{tab:main_math_results}
  \begin{adjustbox}{max width=\textwidth}
  \begin{tabular}{@{}lcccccccc|cc@{}}
  \toprule
  \textbf{Model} & \textbf{AIME24} & \textbf{AIME25} & \textbf{AMC} &
  \textbf{HMMT-F} & \textbf{HMMT-N} & \textbf{Minerva} &
  \textbf{Olympiad} & \textbf{Omega} & \textbf{AVG} & \textbf{$\Delta$} \\
  \midrule
  \multicolumn{11}{@{}l}{\emph{ Reference API models  }} \\
  \midrule
  Claude Opus 4.6   & 98.3 & 68.3 & 97.0 & 52.5 & 61.7 & 60.9 & 88.6 & 45.2 & 71.6 & --- \\
  Gemini 3.1 Pro    & 82.5 & 63.3 & 93.4 & 59.2 & 69.2 & 60.8 & 76.4 & 48.2 & 69.1 & --- \\
  Gemini 3 Flash    & 66.7 & 48.3 & 86.1 & 50.0 & 55.0 & 59.4 & 72.6 & 35.5 & 59.2 & --- \\
  GPT-5.2           & 62.5 & 50.0 & 79.2 & 28.3 & 45.8 & 53.3 & 67.9 & 36.5 & 53.0 & --- \\
  MiniMax M2.7      & 56.7 & 60.0 & 84.9 & 45.0 & 48.3 & 54.1 & 64.7 & 39.0 & 56.6 & --- \\
  \midrule
  \multicolumn{11}{@{}l}{\emph{Base model: \texttt{gpt-oss-120b}}} \\
  \midrule
  \texttt{gpt-oss-120b} (base)
    & 70.0 & 63.3 & 89.0 & 40.0 & 50.0 & 55.1 & 71.4 & 50.8 & 61.2 & --- \\
  \quad + RL (full source mixture)
    & 80.0 & 56.7 & \textbf{93.9} & 43.3 & 46.7 & 52.9 & 72.3 & 52.4 & 62.3 & +1.1 \\
  \quad + RL (Golden Goose~\citep{lu2026golden})
    & 80.0 & 60.0 & 91.5 & 43.3 & 50.0 & 54.8 & 72.0 & 53.2 & 63.1 & +1.9 \\
    \midrule
  \quad + RL (in-budget Goldilocks source subset) 
    & 79.2	& 74.6 &	{93.7}	& 49.2 &	57.7	& 54.5 &	73.6 &	53.5	&	67.0&	+5.8   \\
  \quad + RL (Seed synthesis Goldilocks, synthetic) 
   & 73.3	& 73.1&	92.9&	51.2&	\textbf{60.0}	&54.7&	76.1&	53.2&	  	 66.8&	+5.6   \\
  \rowcolor{bestcell}
  \quad + RL (\textbf{MagicBeanStalk, synthetic})
    & \textbf{80.2} & \textbf{78.8} & {93.4} & \textbf{57.5}
    &  {59.0} & \textbf{55.5} & \textbf{78.2} & \textbf{55.9}
    & \textbf{69.8} & \textbf{+8.6} \\
  \midrule
  \multicolumn{11}{@{}l}{\emph{Base model: DeepSeek~v4~Flash}} \\
  \midrule
  DeepSeek~v4~Flash (base)
    & 76.7 & 80.0 & 92.7 & 73.3 & 76.7 & 57.0 & 75.4 & 47.6 & 72.4 & --- \\
  \rowcolor{bestcell}
  \quad + RL (\textbf{MagicBeanStalk, same data})
    & \textbf{93.3} & \textbf{83.3} & 92.7 & \textbf{83.3} & \textbf{83.3}
    & \textbf{59.2} & \textbf{77.5} & \textbf{50.2} & \textbf{77.9}
    & \textbf{+5.4} \\
  \bottomrule
  \end{tabular}
  \end{adjustbox}
  \vspace{-4mm}
  \end{table*}

  

  \section{Experimental Results}
  \label{sec:experiments}
 In this section, we first quantify qualified yield, blueprint preservation, data quality, and downstream performance relative to the source seeds and seed synthesis in a control experiment. We then evaluate continual RLVR across 8 math benchmarks, transfer to a different target policy, performance beyond the training response budget, and document-grounded finance. Unless specifically mentioned, for most experiments involving training, we use as 16K rollout budget due to computational constrains. 
 
  \subsection{From Zero-Advantage Seeds to Effective RLVR Data}

  \label{sec:exp_data_quality}

  \textbf{Qualified synthesis yield from zero-signal seeds.}
  We randomly sample approximately $1{,}250$ source seeds on which \gptoss
  obtained $0/8$ correct rollouts at 16K. From the same seeds, MBS produces $3{,}370$
  descendants and seed synthesis produces $2{,}494$. A descendant is
  qualified if it lies in the Goldilocks band and has at least one correct
  rollout that terminates normally within 16K. MBS yields qualified descendants
  from $74.7\%$ of the seeds, compared with $43.8\%$ under seed synthesis;
  its candidate-level qualified rate is also higher ($42.8\%$ versus $31.7\%$;
  Table~\ref{tab:fixed_seed_filtering}). Our algorithm  notably converts  more
  zero-signal hard seeds into usable training data.

  \textbf{Blueprint preservation and novelty.}
  For each method, GPT-5.6 scores $300$ math and $100$ finance seed--candidate
  pairs from 1 to 5 for blueprint preservation and substantive novelty. In
  math, our method preserves the blueprint comparably to seed synthesis while
  producing more novel problems. Concretely, $19.7\%$ of MBS pairs score at least 4 on both
  criteria, compared with $3.0\%$ for seed synthesis. In finance, the
  corresponding rates are $86.0\%$ and $17.0\%$
  (Figure~\ref{fig:seed_synthesis_analysis}, left).

  \textbf{Data quality analysis.}
  Three frontier models  (GPT-5.6, Opus 4.8 and Gemini 3.1 Pro)  assess whether each of $150$ math and
  $50$ finance randomly sampled problems are well-posed and its GT answer is correct. For 
  finance,we also include the source document. 
   All three judges reject only $8.0\%$ of the math answers and none of the finance answers (Table~\ref{tab:mbs_gt_quality}). We report all agreement metrics but emphasize the usage of unanimous rejection. Failure to verify an answer may also reflect the judge’s limitations on difficult problems. Our hard-seed control experiment (\ref{app:hard_seeds_frontier}) shows this.

 \textbf{Downstream performance analysis.}
Starting from the same \gptoss checkpoint, we analyze the training dynamics
for one epoch on data derived from the same fixed seed population, and report
the average over the 4 core benchmarks (AIME~24/25, HMMT-F/N). The 
average is $65.3$ for source-seed training, $63.9$ for Goldilocks-filtered
seed synthesis, and $69.2$ for MBS
(Figure~\ref{fig:seed_synthesis_training_main}, left). Seed synthesis does not
improve over the source seeds, whereas MBS improves over both. This is a
separate control run and is no the full-budget 
results in Table~\ref{tab:main_math_results}.

  \textbf{Performance on the original hard seeds after synthetic training.}
  We reevaluate approximately $1{,}250$ source seeds on which the initial
  \gptoss sample obtained $0/8$ at 16K. A fresh base sample recovers $9.5\%$ of
  these seeds, compared with $23.2\%$ after seed synthesis training and
  $25.0\%$ after MagicBeanStalk training. MagicBeanStalk also yields $840$ correct rollouts and
  masters $14$ seeds, compared with $694$ and $6$ under seed synthesis
  (Table~\ref{tab:hard_seed_recovery}). Both methods transfer learning back to
  their source seeds, while MBS produces more correct rollouts and more fully
  mastered seeds.

  \begin{table}[t]
  \centering
  \caption{\textbf{Inference-budget evaluation of 16K-trained checkpoints.}
  Parentheses report the difference from the base policy at the same inference
  budget. \emph{Top:} average pass@1 across the 8 math benchmarks, average over 3 samples. \emph{Bottom:} pass@1 on MathArena Apex-Shortlist ($47$
  problems), average over 16 samples per problem, evaluated from the 16K training budget
  upward.}
  \label{tab:inference_budget_generalization}
  \small
  \setlength{\tabcolsep}{2.8pt}
  \begin{adjustbox}{max width=\columnwidth}
  \begin{tabular}{@{}lcccc@{}}
  \toprule
  \textbf{Training condition} & \textbf{8K} & \textbf{16K} & \textbf{24K} & \textbf{32K} \\
  \midrule
  \multicolumn{5}{@{}l}{\emph{8 math benchmarks}} \\
  \gptoss{} (base)
    & $49.7$ & $61.3$ & $72.5$ & $75.9$ \\
  \quad + RL (in-budget Goldilocks source subset)
    & $54.0$ {\scriptsize $(+4.3)$} & $67.1$ {\scriptsize $(+5.8)$}
    & $69.6$ {\scriptsize $(-2.9)$} & $72.0$ {\scriptsize $(-3.9)$} \\
  \quad + RL (Seed synthesis, Goldilocks synthetic)
    & $54.2$ {\scriptsize $(+4.5)$} & $66.4$ {\scriptsize $(+5.1)$}
    & $72.5$ {\scriptsize $(\ \ 0.0)$} & $75.6$ {\scriptsize $(-0.3)$} \\
  \rowcolor{bestcell}
  \quad + RL (\textbf{MagicBeanStalk, synthetic})
    & $\mathbf{58.3}$ {\scriptsize $(\mathbf{+8.6})$} & $\mathbf{69.6}$ {\scriptsize $(\mathbf{+8.3})$}
    & $\mathbf{73.5}$ {\scriptsize $(\mathbf{+1.0})$} & $\mathbf{77.4}$ {\scriptsize $(\mathbf{+1.5})$} \\
  \midrule
  \multicolumn{5}{@{}l}{\emph{MathArena Apex-Shortlist}} \\
  \gptoss{} (base)
    & -- & $1.5$ & $6.8$ & $16.2$ \\
  \quad + RL (in-budget Goldilocks source subset)
    & -- & $7.4$ {\scriptsize $(+5.9)$} & $15.0$ {\scriptsize $(+8.2)$}
    & $18.9$ {\scriptsize $(+2.7)$} \\
  \quad + RL (Seed synthesis, Goldilocks synthetic)
    & -- & $4.7$ {\scriptsize $(+3.2)$} & $12.9$ {\scriptsize $(+6.1)$}
    & $20.6$ {\scriptsize $(+4.4)$} \\
  \rowcolor{bestcell}
  \quad + RL (\textbf{MagicBeanStalk, synthetic})
    & -- & $\mathbf{12.5}$ {\scriptsize $(\mathbf{+11.0})$} & $\mathbf{20.1}$ {\scriptsize $(\mathbf{+13.3})$}
    & $\mathbf{27.1}$ {\scriptsize $(\mathbf{+10.9})$} \\
  \bottomrule
  \end{tabular}
  \end{adjustbox}
  \vspace{-4mm}
\end{table}

  \subsection{Continual RLVR on Math}
  \label{sec:exp_on_policy}
  \textbf{Training data and baselines.}
  We draw math seeds from an approximately $78$K-problem open-source mixture of
  DAPO-Math-17K~\citep{Yu2025DAPO},
  Skywork-OR1-RL-Data~\citep{he2025skywork}, Turing
  Open-RL~\citep{turing_2026_open_rl}, and a subset of StackMathQA~\citep{zhang2024stackmathqa}.  We
  compare MBS with the full source mixture, its in-budget Goldilocks subset,
  Goldilocks-filtered seed synthesis, and Golden
  Goose~\citep{lu2026golden} . Our work focus is on the effectiveness of the synthesized data rather than to minimize synthesis cost or algorithm hyperparams. Therefore, training set sizes are each subset Goldilocks counts(Fig~\ref{fig:difficulty_controls_main}a) refiltered to match 16K training rollouts . RL
  implementation and  hypers are fixed. We
  stop when a decline is not recovered within 10 subsequent iterations on the same held-out synthetic subset.

  \textbf{Training setup.}  We train \gptoss with GRPO on 2 nodes (16 H100 GPUs), using LoRA (rank 32), a constant lr of $4\times10^{-5}$, and a 16K response limit. Each
  step samples 16 responses for each of 32 prompts  , with an optimization batch size of 32. DeepSeek~v4~Flash uses full-model GRPO on 10 nodes (80 H100 GPUs),
  with a $10^{-6}$ lr and 8 responses  under the same 16K limit. We use miles with Megatron-LM training and SGLang inference engine. To reject sample correct solutions, we use a pool of teachers including: Gemini 3 Flash, 
Claude Opus 4.6, DeepSeek V4 Pro, GPT 5.2 and Kimi K2.6. We use the same pool for both seed synthesis and MBS.

  \textbf{Evaluation.}
  We report pass@1 with a 16K limit on the complete 8-benchmark suite:
  AIME~2024/2025, AMC, HMMT February and November, Minerva, Olympiad, and
  Omega-500. \gptoss uses 16
  rollouts per problem. API rows use 4. Answers are checked using 
  normalized string matching , numeric equivalence and GPT-5.2 as a judge.

  \textbf{Main result.}
  Continual RLVR on MBS data raises the \gptoss average from $61.2$ to $69.8$
  ($+8.6$ points) and improves most benchmarks
  (Table~\ref{tab:main_math_results}). Training on the full source
mixture reaches $62.3$ and on Golden Goose $63.1$, while the in-budget
Goldilocks source subset and seed synthesis reach $67.0$ and $66.8$.
We use the released Golden Goose corpus as-is, without regenerating or
re-filtering it for \gptoss, so this row is a published-data reference
rather than a re-run of that method under our target policy.
\textbf{Transfer to a different target policy.}
We train DeepSeek~v4~Flash on the same MBS data,  without
  regeneration or target-specific re-filtering, and evaluate with 1 rollout per
  problem at 16K. Its 8-benchmark average rises from $72.4$ to $77.9$ ($+5.4$ points.  Table~\ref{tab:main_math_results}).

  \begin{table}[t]
  \begin{minipage}[t]{0.64\linewidth}
  \vspace{0pt}
  
  \centering
  \caption{\textbf{Continual RLVR on MagicBeanStalk financial data.}
  Pass@1 and pass@5 (\%) on SecQue and FinanceBench using 8 rollouts and an
  8K maximum response length. MBS improves on both benchmarks.}
  \label{tab:finance_results}
  \small
  \resizebox{\columnwidth}{!}{%
  \begin{tabular}{@{}lrrrr@{}}
  \toprule
   & \multicolumn{2}{c}{\textbf{SecQue}} &
     \multicolumn{2}{c}{\textbf{FinanceBench}} \\
  \cmidrule(lr){2-3} \cmidrule(lr){4-5}
  \textbf{Model} & \textbf{Pass@1} & \textbf{Pass@5} &
  \textbf{Pass@1} & \textbf{Pass@5} \\
  \midrule
  Qwen3-4B-Instruct         & 32.4 & 43.1 & 65.9 & 76.0 \\
  \rowcolor{bestcell}
 + RL (\textbf{MagicBeanStalk})
                            & \textbf{39.3} & \textbf{52.5} &
                              \textbf{68.3} & \textbf{76.8} \\
  \bottomrule
  \end{tabular}
  }
  \end{minipage}
  \begin{minipage}[t]{0.35\linewidth}
  \vspace{0pt}
  \caption{\textbf{Judge-assessed quality of our data.}
   Three frontier  judges classify 
  synthetic problem--answer
  as correct, incorrect, or uncertain.
  }
  \label{tab:mbs_gt_quality}
  
  \centering
  \scriptsize
  \setlength{\tabcolsep}{3.0pt}
  \resizebox{\columnwidth}{!}{%
  \begin{tabular}{@{}lrrr@{}}
  \toprule
  \textbf{Domain} &
  \shortstack{\textbf{At least one}\\\textbf{verifies}} &
  \shortstack{\textbf{Majority}\\\textbf{verifies}} &
  \shortstack{\textbf{All 3}\\\textbf{reject}} \\
  \midrule
  Math    & $89.3\%$  & $80.0\%$ & $8.0\%$ \\
  Finance & $100.0\%$ & $80.0\%$ & $0.0\%$ \\
  \bottomrule
  \end{tabular}
  }
  \end{minipage}
  \vspace{-4mm}
  \end{table}

  \textbf{Does the improvement persist beyond the training response
  budget?}
  In this experiment,  we hold each checkpoint and the evaluation protocol fixed
  and vary only the inference response length. All 3 trained policies improve
  over the base policy at 16K. At 24K and 32K, however, source-seed training
  falls below the base policy and seed synthesis converges to it, whereas MBS
  remains above base by $1.0$ and $1.5$ points, respectively
   (Table~\ref{tab:inference_budget_generalization}, top). Next, we evaluate the same
  checkpoints at the same budgets on a substantially harder set known to benefit
  from larger responses. MathArena Apex-Shortlist \citep{matharena_apex}
  contains $47$ competition problems. MBS improves over base by
  $11.0$, $13.3$, and $10.9$ points at 16K, 24K, and 32K
  (Table~\ref{tab:inference_budget_generalization}, bottom).

  \begin{table}[t]
   \vspace{-2mm}
  \centering
  
  \begin{minipage}[t]{0.49\linewidth}
  \vspace{0pt}
  \caption{\textbf{Qualified yield from hard seeds.}
  We synthesize descendants from approximately $1{,}250$ seeds on which
  \gptoss obtained $0/8$ correct rollouts at 16K. A qualified descendant is
  Goldilocks and admits a normally terminated correct rollout; seed conversion
  measures whether a seed yields at least one such descendant. MBS improves
  both candidate-level yield and seed-level conversion, turning substantially
  more hard seeds into usable RLVR data.}
  \label{tab:fixed_seed_filtering}
  
  \centering
  
  \setlength{\tabcolsep}{2.5pt}
  \resizebox{\columnwidth}{!}{%
  \begin{tabular}{@{}lrrrr@{}}
  \toprule
  \textbf{Method} &
  \textbf{Desc.} &
  \shortstack{\textbf{Goldilocks}\\\textbf{rate}} &
  \shortstack{\textbf{Qualified}\\\textbf{rate}} &
  \shortstack{\textbf{Seed}\\\textbf{conversion}} \\
  \midrule
  Seed synthesis & $2{,}494$ & $36.8\%$ & $31.7\%$ & $43.8\%$ \\
  MagicBeanStalk              & $3{,}370$ & $55.4\%$ & $42.8\%$ & $74.7\%$ \\
  \bottomrule
  \end{tabular}
  }
  \end{minipage}
  \hfill
  \begin{minipage}[t]{0.49\linewidth}
  \vspace{0pt}
  
  \centering
  \caption{\textbf{Recovery on the original hard seeds.}
    Evaluation of the
     models trained with seed
    synthesis and MBS on approx $1{,}250$ source seeds that received $0/8$ at 16K. Recovery requires at lest one correct rollout; correct rollouts aggregate all attempts. Both 
    strategies transfer
    to the original hard seeds. This is
    expected for seed synthesis as
    variations effectively augments the seed problem neighborhood. MBS
    preserves transfer,
    master
    more source seeds, and generalizes better.
    }

  \label{tab:hard_seed_recovery}
  \scriptsize
  \setlength{\tabcolsep}{3.2pt}
  \resizebox{\columnwidth}{!}{%
  \begin{tabular}{@{}lccc@{}}
  \toprule
  \textbf{Policy} & \textbf{Recovered $\geq 1/8$} &
  \shortstack{\textbf{Correct rollouts}} &
  \textbf{Mastered $8/8$} \\
  \midrule
  Fresh base control       & $118\;(9.5\%)$ & $164\;(1.7\%)$ & $0\;(0.0\%)$ \\
  Seed Synthesis & $288\;(23.2\%)$ & $694\;(7.0\%)$ & $6\;(0.5\%)$ \\
  MBS              & $310\;(25.0\%)$ & $840\;(8.5\%)$ & $14\;(1.1\%)$ \\
  \bottomrule
  \end{tabular}
  }
  \end{minipage}
  \end{table}

  \subsection{MagicBeanStalk for Finance Data}
  \label{sec:exp_finance}
%
  \textbf{Training data.}
  Starting from  $340$ expert-collected questions grounded in 17 corporate credit agreements, MBS generates approximately $25$K
  document-grounded candidates from the same seeds. MBS yields $14.2$K Goldilocks problems (Figures~\ref{fig:difficulty_controls_main}.
 
  \textbf{Training setup.}
  We use GRPO in veRL to train Qwen3-4B-Instruct on the MBS corpus with an 8K response limit. We use a batch size of 512 prompts, 16 responses per prompt,  and a lr of $10^{-6}$. Two additional nodes host Qwen3-30B-A3B-Instruct-2507, which assigns binary rewards by comparing each extracted final answer with the reference under a finance-aware equivalence rubric.
  \textbf{Evaluation.}
  We evaluate the Qwen3-4B-Instruct  , and MBS-trained policy on SecQue~\citep{benyoash2025secque} and FinanceBench~\citep{islam2023financebench}. We follow each benchmarks eval protocol, sample 8 responses per problem at
  temperature $1.0$ with an 8K response limit and report pass@1 and pass@5 and use GPT-5.2 as a judge. 
  
\textbf{Main result.}
  Table~\ref{tab:finance_results} reports continual RLVR of Qwen3-4B-Instruct
  on the MBS finance corpus. Training on MBS improves pass@1  by \textbf{$6.9$} points on SecQue and \textbf{$2.4$} points on FinanceBench.

\section{Related work}

\textbf{Scaling Training Data for RLVR.}
Collected corpora supply verified math and code problems~\citep{albalak2025bigmathlargescalehighqualitymath,gao2024omnimathuniversalolympiadlevel,chen2025acereason,deepscaler2025,cui2025process,lu2025scp116khighqualityproblemsolutiondataset,jain2024livecodebenchholisticcontaminationfree,liu2025rstarcoderscalingcompetitivecode,NEURIPS2025_82a34882,lin2026goedel,lu2504retro}. Others generate tasks in hand-designed environments~\citep{lacombe2025reasoning,stojanovski2025reasoninggymreasoningenvironments}, adapt difficulty to the policy~\citep{Zeng2025RLVESU,xu2026scaler}, or mine tasks from Internet text~\citep{lu2026golden}. Closest to us,
a teacher writes new problems from seeds or concepts for SFT~\citep{ toshniwal2024openmathinstruct2,zhao2025promptcot}, VLMs~\citep{liao2025longperceptualthoughts,acuna2025long,jung2026deltaprompts,acuna2025socratic}, and RLVR~\citep{yu2025cot,liang2025swsselfawareweaknessdrivenproblem,liang2026beyond}. 
Rather than collecting problems, hand-engineering environments, scaling quantity,  or asking a teacher to synthesize  based on seeds, we analyze what makes a synthetic problem useful for RLVR, then design a top-down, bottom-up recursive algorithm that turns hard seeds into  new, policy- and budget-calibrated problems.

\textbf{Algorithmic RL for Hard Problems.}
Dynamic sampling discards zero-advantage groups~\citep{Yu2025DAPO}, BroRL scales rollouts per prompt~\citep{hu2025brorlscalingreinforcementlearning}. Recent work guide exploration with oracle prefixes, hints, or off-policy traces~\citep{qu2026pope,li2026questa,zhang2026scaf}. These methods modify the training algorithm or optimizer. Our work is complementary and focuses on the data rather than the modifying the algorithm with the same problems.

%

\section{Conclusion}
We show the value of RLVR data is relative to the problem, target policy, and rollout budget. 
We also show
hard problems should guide the type of problems to synthesize. 
Based on these findings, 
we design key principles to guide effective data synthesis algorithms. 
MBS instantiates these principles by decomposing hard seeds into validated reasoning trees and regrowing them into substantively new, policy- and budget-calibrated problems. Our data improves \gptoss by $+8.6$ and DeepSeek~v4~Flash by $+5.4$ points across 8 math benchmarks. In finance, $340$ expert questions yield over $14{,}000$ verifiable training problems and improve
Qwen3 by  +6.9  and +2.4  points 
on SecQue and FinanceBench, respectively.

\subsection*{AI use statement}

We used generative AI tools to help with language polishing and improve clarity. We additionally use LLM to generate synthetic datasets as this is the core of our work. The AI tools  did not contribute to the intellectual content. We  reviewed all AI-assisted edits and take responsibility for the final content of this work.

\subsection*{Reproducibility statement}

We use publicly available models and benchmarks with dataset splits and training hyperparameters,  implementation details, and checkpoint selection disclosed in Section~\ref{sec:experiments}. Prompt templates for synthesis and evaluation appears in Section~\ref{app:core-prompt-templates} and~\ref{app:judge-analysis-protocols}. Other reproducibility details are in Section~\ref{app:reproducibility-details}. Reported results are averaged over multiple rounds as disclosed in in Section~\ref{sec:experiments}.   One caveat may limit exact reproduction of the synthesis algorithm from scratch. Some of teachers used to reject sample solutions are proprietary models served behind APIs whose updates we do not control, so solutions and problems synthesized using them may differ if  APIs change. We will the release the generated MagicBeanStalk datasets upon acceptance.

\subsection{Ethics Statement}
We use LLM to generate training problems, so the resulting data may contain errors. We audit a sample with three additional models and report their agreement and rejection rates, while recognizing that these judgments may also contain mistakes. Our finance experiments evaluate benchmark performance and are intended to showcase the strength of the proposed algorithms.  Further testing would be needed before relying on synthetic data to train models for financial decisions.

\bibliographystyle{iclr2027_conference}

\bibliography{reference}

\newpage
\appendix
\newpage
\section{Appendix}

\section{What Makes a Problem Useful for RLVR?}
\label{app:useful_rlvr_data}

\subsection{A large corpus can provide little learning signal}
\label{app:abundant_little_signal}

We next use the empirical success rate to characterize the learning signal
available in a large corpus of open-source mathematical reasoning problems.
The corpus combines {problems drawn from
  DAPO-Math-17K~\citep{Yu2025DAPO},
  Skywork-OR1-RL-Data~\citep{he2025skywork}, Turing
  Open-RL~\citep{turing_2026_open_rl}, and a subset of StackMathQA~\cite{zhang2024stackmathqa},
} and contains
approximately $78{,}000$ problems. For each problem, we sample $N=8$ responses
from \gptoss using rollout budget $L$.
Despite the size of the corpus, only $12.2\%$ of its problems fall within the
observed Goldilocks band ($L=32K$). As shown in Figure~\ref{fig:difficulty_controls_main},
most problems are solved in all rollouts and therefore have
empirical success rate $\bar r_{\pi,L}(\problem)=1$ while a $6.3\%$ are not solved in any rollout. 
Therefore, roughly \textbf{$\mathbf{88\%}$ of the corpus falls in the zero-advantage trap}. They contribute  no group-relative advantage.

We then test whether the empirical Goldilocks criterion identifies problems
that are more useful for downstream RLVR. Starting from the same initial
\gptoss policy, we compare three computationally tractable training sets. We
first construct a $\sim$7K-problem proxy whose proportions of hard,
Goldilocks, and stale problems match those of the full open-source corpus.
From this proxy, we extract the $\sim$862 Goldilocks problems and a stricter
696-problem subset that also demonstrates in-budget solvability under the 16K
training rollout limit. All
conditions use the same training recipe and are evaluated after approximately
one epoch.
As shown in Figure~\ref{fig:seed_synthesis_training_main}(b), both the
Goldilocks-only and the Goldilocks plus in-budget subsets reach a higher
average accuracy than the full-mixture proxy, while training on an order of
magnitude fewer problems.
Thus, Goldilocks filtering concentrates useful RLVR signal.
Filtering, however, can only select. It retains the small fraction of a corpus
that already yields mixed outcomes and discards the rest, including the hard
problems on which the policy never succeeds.
Most importantly,  filtering cannot produce a problem
for a capability the corpus does not already cover.

\subsection{Difficulty is relative to the policy and training budget}
\label{app:difficulty_training_budget}

Problem difficulty in RLVR 
depends on both the target policy and the rollout budget used during training.
We demonstrate each dependence through a controlled experiment. Let $f_i\in\{\texttt{stop},\texttt{length}\}$ indicate whether response
terminates normally or reaches the rollout limit $L$. We say
that a problem has demonstrated \emph{in-budget solvability} if at least one
correct response terminates:
\begin{equation}
Q_{\pi,L}(\problem)=
\mathbf{1}\!\left[
\exists i:\ r_i=1
\land f_i=\texttt{stop}
\right].
\label{eq:in_budget_solvability}
\end{equation}

\paragraph{Dependence on the target policy.}
We first hold the problem set and rollout budget fixed and vary only the
policy. We use the $15{,}470$ problems that GPT-5-nano does not solve in every
rollout, discarding those it has already mastered, and evaluate them also  with \gptoss using 8 rollouts per problem and 32K response limit.
Changing only the policy causes nearly one third of the problems ($32.3\%$)
to move between the zero-success, Goldilocks, and fully solved regions. Thus relative to a target policy, the same problem can provide different
learning signal.

\paragraph{Dependence on the rollout budget.}
We next hold the policy and evaluation set fixed and vary only the maximum
response length. 
Increasing the rollout budget of \gptoss from 16K to 64K reduces the fraction of
hard problems from $33.1\%$ to $10.6\%$ and raises the average accuracy on the
8 math benchmarks from $61.2\%$ to $78.9\%$
(Figure~\ref{fig:difficulty_controls_main}b).
Thus, the same problems appear substantially
harder under a smaller rollout budget, even though the policy is unchanged.
Therefore, a
truncated response does not establish that the policy lacks the reasoning
capability required by the problem. It only establishes  that no complete
solution was produced within budget.
We therefore treat a problem as providing a clean in-budget group-relative
learning signal only when:
\begin{equation}
0<\bar r_{\pi,L}(\problem)<1
\qquad\text{and}\qquad
Q_{\pi,L}(\problem)=1.
\label{eq:clean_in_budget_signal}
\end{equation}

Difficulty is consequently a property of the tuple
$(\problem,\pi,L)$, rather than of the problem alone.

\subsection{Verified Hard problems specify frontier capabilities}

\label{app:hard_seeds_frontier}
Hard problems play a paradoxical role in RLVR. Problems with empirical success
rate zero provide no direct group-relative advantage signal. Yet, these
problems may still reveal reasoning capabilities that the target policy lacks.
We argue that a verified solution trace exposing the essential
reasoning operations together with how those operations
are composed reveals the capability the target policy lack. We refer to this structure as the problem's
\emph{reasoning blueprint}. It provides a specification of capabilities at the
frontier of the current policy.
We define a \emph{verified hard problem} as a solvable/correct problem for which
$\bar r_{\pi,L}(\problem)=0$ and all sampled responses terminate normally:
\begin{equation}
\bar r_{\pi,L}(\problem)=0
\qquad\text{and}\qquad
f_i=\texttt{stop}
\quad \forall i\in\{1,\ldots,N\}.
\label{eq:verified_hard}
\end{equation}
This termination requirement ensures that the observed failures reflect a
reasoning limitation rather than an artifact of the rollout budget.

\paragraph{Extracting reasoning blueprints.}
We study $98$ math hard seed problems randomly selected from the open corpus. For each seed, we obtain a verified solution and reasoning trace from a pool of frontier models. We use a pool of teacher ( Gemini 3 Flash, 
Claude Opus 4.6, DeepSeek V4 Pro, GPT 5.2 and Kimi K2.6) as we found  with one  attempt per model, the strongest individual teacher solved  only (59/98) seeds.  From each trace, we then extract
the essential reasoning operations and how they are composed (i.e. we extract their \emph{reasoning blueprint}).

\paragraph{Controlled in-context experiment.}
For each seed, we generate a simpler subproblem  that instantiates some of the operations in the reasoning blueprint and provide its verified solution as a single
in-context example to the target policy. We use an LLM as a judge to verify the subproblem does not explicitly reveal the solution of the original problem. For simplicity, and to avoid best-of-$K$ selection effects, we use only the first generated subproblem. After this intervention, the
policy solves at least one rollout for $78/98$ seeds ($79.6\%$), while mean
accuracy on the original hard problems increases from $0\%$ to $46.9\%$.
These results   suggest that hard problems   provide no direct RLVR signal
while still serving as useful specifications of missing reasoning capabilities.

\paragraph{Are blueprints consistent across frontier models?}
Among the $43$ seeds with at least two verified
blueprints, we find $90.7\%$ produce the same reasoning blueprint or a
compatible variant. This suggests that the blueprint primarily reflects the
structure of the problem rather than the idiosyncrasies of a particular
frontier model.

\paragraph{Observation.}
Hard problems are not merely training examples; they can be specifications of
capabilities at the frontier of the current policy. We call a verified problem
a \emph{capability-frontier seed} when the target policy cannot solve it
reliably and its solution specifies a reasoning capability, or composition of
capabilities, beyond the policy's current competence. 
The in-context experiment
shows one of the many actionable insights: a simpler, subproblem
requiring a capability observed in the blueprint of the original problems solution can substantially improves the target policy's success on the hard problem.

\section{Seed Synthesis: Why Direct Seed Synthesis Is Not Enough}
\label{app:direct_seed_synthesis}

Filtering can only select from problems a corpus already contains
, while hard seeds specify the
reasoning the policy is missing and left out during filtering. The natural next step is therefore to generate new (easier) problems directly from those
hard seeds. A simple way to synthesize problem starting with hard seeds is
to provide the problem, reasoning trajectory, and its verified solution to a strong teacher
%
, and perform rejection sampling on quality control, goldilocks and in-budget solvability. If $(\problem,\tau,a^*)$ denotes the seed problem, reasoning trace, and verified answer, seed synthesis samples:
\begin{equation}
  (\widetilde \problem,\widetilde\tau,\widetilde a^*)
  \sim T(\problem,\tau,a^*).
  \label{eq:direct_seed_synthesis}
\end{equation}
This baseline is straightforward and we experimentally observed it often retains the mathematical techniques used by the seed. In a controlled set of experiments, however, we observed \textbf{seed synthesis provides no clear gain beyond the seeds.} Starting from the same base policy and the same fixed seed population, we compare training on the $2{,}195$ source seeds with training on approximately the same number of Goldilocks-filtered synthetic problems derived from those seeds. Figure~\ref{fig:seed_synthesis_training_main}(a) shows the training curves. The best average accuracy across the 4 core math benchmarks (AIME~24/25, HMMT-F/N) is $65.3$ for seed-only training and $63.9$ for seed synthesis.
Therefore, this simple method produces additional problem instances, but does not provide a clear downstream improvement in unseen benchmarks. We then investigate why:

\paragraph{Seed synthesis preserves techniques but often makes shallow
variants.}
We study this behavior with a blinded judge study over $300$ generated
problems in math and $100$ in finance. The judge takes in  the seed, the
reasoning blueprint and one generated problem with its reference
solution, but not the synthesis method. It evaluates whether the generated
solution requires the operations and dependencies in the seed blueprint, and
whether the problem is substantively new rather than a paraphrase, contextual
rewrite, or numerical substitution. Figure~\ref{fig:seed_synthesis_analysis} shows that
seed synthesis often retains recognizable reasoning techniques, particularly
in math, but frequently does so through a shallow variation of the
original problem rather than a substantively new instantiation .

\paragraph{Post-hoc filtering calibrates difficulty but cannot repair the
candidate pool.}
Filtering is subtractive. It can calibrate the difficulty and response length
of the candidates it retains, but it cannot turn a shallow variant into a
substantively new problem, nor produce a candidate for a seed that yielded
none. Seed synthesis consequently returns below one qualified problem per seed
(Figure~\ref{fig:seed_synthesis_analysis}, right). We study the full yield and
seed-coverage comparison in Section~\ref{sec:exp_data_quality}.

\textbf{What changes when direct synthesis receives the MBS inference
  budget?} We compare direct synthesis and MBS on the same 64 seeds. For the matched control, we repeatedly generate and verify direct-synthesis candidates
  from each seed until reaching or exceeding MBS's per-seed construction budget. The resulting control receives 16.0M estimated construction tokens, compared
  with MBS's 14.3M reference budget. We evaluate blueprint preservation and
  novelty using the same  protocol as above, and estimate
  Goldilocks difficulty from $N=8$ \gptoss rollouts with a 16K limit. As shown in Table~\ref{tab:ablation_resource_and_yield}, additional sampling improves direct synthesis's Goldilocks coverage, but does not close its novelty gap. MBS yields an all-three candidate for 28 seeds, compared with 13 under the matched direct-synthesis control.

  \begin{table*}[h]
    \centering
    \caption{Direct synthesis under the MBS construction budget. Samples is the number of evaluated candidates. Blueprint and
    novelty require judge scores of at least three, and Goldilocks denotes
    $1$--$7$ correct responses among $N=8$ \gptoss rollouts at 16K. }
    \label{tab:ablation_resource_and_yield}
    \small
    \setlength{\tabcolsep}{5.0pt}

    \begin{tabular}{@{}lrrrrr@{}}
      \toprule
      Method & Samples & Blueprint & Novelty & Goldilocks & All three \\
      \midrule
      Direct synthesis & 64 & $48/64$ & $17/64$ & $32/64$ & $3/64$ \\
      Direct synthesis (oversampling) & 387 & $61/64$ & $29/64$ & $56/64$ & $13/64$ \\
      MBS & 64 & $52/64$ & $45/64$ & $48/64$ & $\mathbf{28/64}$ \\
      \bottomrule
    \end{tabular}
  \end{table*}

\paragraph{Observation.}
With todays frontier models, a single call from the hard seed does not produce enough valid, new, blueprint-preserving problems for subsequent policy- and
budget-based filtering.

Together with the preceding analysis of policy- and budget-relative learning
signal, this result suggests a simple separation of roles. Hard problems
specify the reasoning to target; the target policy and training rollout budget
determine which generated problems are useful for RLVR. Ideally, we should aim to generate substantively
new problems that preserve the hard seed's essential reasoning operations and
their composition, then retain those that yield both successful and
unsuccessful target-policy rollouts and at least one correct
target-policy rollout within the training budget.

  \begin{table}[t]
  \centering
  
  \begin{minipage}[t]{0.49\linewidth}
  \vspace{0pt}
  \caption{\textbf{Qualified yield from hard seeds.}
  We synthesize descendants from approximately $1{,}250$ seeds on which
  \gptoss obtained $0/8$ correct rollouts at 16K. A qualified descendant is
  Goldilocks and admits a normally terminated correct rollout; seed conversion
  measures whether a seed yields at least one such descendant. MBS improves
  both candidate-level yield and seed-level conversion, turning substantially
  more hard seeds into usable RLVR data.}
  \label{tab:fixed_seed_filtering}
  
  \centering
  
  \setlength{\tabcolsep}{2.5pt}
  \resizebox{\columnwidth}{!}{%
  \begin{tabular}{@{}lrrrr@{}}
  \toprule
  \textbf{Method} &
  \textbf{Desc.} &
  \shortstack{\textbf{Goldilocks}\\\textbf{rate}} &
  \shortstack{\textbf{Qualified}\\\textbf{rate}} &
  \shortstack{\textbf{Seed}\\\textbf{conversion}} \\
  \midrule
  Seed synthesis & $2{,}494$ & $36.8\%$ & $31.7\%$ & $43.8\%$ \\
  MagicBeanStalk              & $3{,}370$ & $55.4\%$ & $42.8\%$ & $74.7\%$ \\
  \bottomrule
  \end{tabular}
  }
  \end{minipage}
  \hfill
  \begin{minipage}[t]{0.49\linewidth}
  \vspace{0pt}
  
  \centering
  \caption{\textbf{Recovery on the original hard seeds.}
    Evaluation of the
    base policy and models trained with seed
    synthesis and MBS on approx $1{,}250$ source seeds that received $0/8$ at 16K. Recovery requires at lest one correct rollout; correct rollouts aggregate all attempts. Both 
    strategies transfer
    to the original hard seeds. This is
    expected for seed synthesis as
    variations effectively augments the seed problem neighborhood. MBS
    preserves transfer,
    master
    more source seeds, and generalizes better to math benchmarks (see main table).}
  
  \label{tab:hard_seed_recovery}
  \scriptsize
  \setlength{\tabcolsep}{3.2pt}
  \resizebox{\columnwidth}{!}{%
  \begin{tabular}{@{}lccc@{}}
  \toprule
  \textbf{Policy} & \textbf{Recovered $\geq 1/8$} &
  \shortstack{\textbf{Correct rollouts}} &
  \textbf{Mastered $8/8$} \\
  \midrule
  Fresh base control       & $118\;(9.5\%)$ & $164\;(1.7\%)$ & $0\;(0.0\%)$ \\
  Seed Synthesis & $288\;(23.2\%)$ & $694\;(7.0\%)$ & $6\;(0.5\%)$ \\
  MBS              & $310\;(25.0\%)$ & $840\;(8.5\%)$ & $14\;(1.1\%)$ \\
  \bottomrule
  \end{tabular}
  }
  \end{minipage}
  \end{table} 

  \section{Supplementary Reproducibility Details}
\label{app:reproducibility-details}

\paragraph{Teacher, student, and target-policy roles.}
The teacher pool solves seed and synthetic problems, proposes decompositions,
generates new problems, and supplies candidates for verification.  The ordered
student pool $\mathcal{S}$ is used for the decomposition \emph{effectivity}
test.  For an evaluated student $S$, we compare rollouts on a parent problem
with rollouts obtained after conditioning on the verified child problems and
their worked solutions; a decomposition is effective when this conditioning
raises the empirical accuracy of at least one evaluated student.  This test
decides whether the proposed children are useful scaffolds.  It does not decide
whether a final synthetic problem is in the Goldilocks band.

The target policy $\model$ has a separate role.  Final Goldilocks membership
and in-budget solvability are determined only from target-policy rollouts at
the training response limit.  In the reported setup, a retained problem has
$1$--$7$ correct responses among $8$ target-policy rollouts and at least one
correct response that terminates normally within the response budget.  The
target policy can also be a member of $\mathcal{S}$, but a positive effectivity
test from any student never substitutes for this final target-policy screen.
 
\paragraph{Document-grounded finance.}
For finance, the document is part of the problem:
$\problem=(\question,\doc)$.  It is not auxiliary metadata that may be dropped
after synthesis.  Decomposition and verification use the supplied snippet,
and a re-grounded tree uses a new snippet throughout.  Every synthesized
question and its answer must be supported by that snippet alone; facts from
the seed document cannot be carried into the new problem unless they also
appear in the new document.

\section{Core Prompt Templates}
\label{app:core-prompt-templates}

The following are condensed task-defining templates.  Bracketed text denotes
runtime fields.  Production wrappers also enforce serialization and answer
formatting, while generic solving, retry, solution-rewriting, and parsing
instructions are omitted here.  A ``verified solution'' is teacher-side
construction context, not an RLVR training target.

\begingroup
\tcbset{apprompt/.style={
  colback=black!2,
  colframe=black!35,
  boxrule=0.3pt,
  arc=0.6mm,
  left=4pt,
  right=4pt,
  top=3pt,
  bottom=3pt,
  before skip=3pt,
  after skip=5pt
}}

\noindent\textbf{Problem decomposition.}\par
\begin{tcolorbox}[apprompt]
\small
\emph{Input:} \texttt{[PARENT PROBLEM]}, \texttt{[VERIFIED TEACHER
SOLUTION]}, \texttt{[GROUND-TRUTH ANSWER]}, and, when applicable,
\texttt{[DOCUMENT]}.

Identify the key skills and knowledge required by the parent.  Produce two or
three easier, independent, self-contained subproblems that collectively teach
those skills.  Give each subproblem a correct worked solution and one concise
final answer.  If a document is supplied, every subproblem must be answerable
using only that document.
\end{tcolorbox}

\noindent\textbf{Conditioned solving for the effectivity test.}\par
\begin{tcolorbox}[apprompt]
\small
\emph{Input:} \texttt{[VERIFIED CHILD PROBLEMS AND WORKED SOLUTIONS]},
\texttt{[PARENT PROBLEM]}, and, when applicable, \texttt{[DOCUMENT]}.

Study the solved child problems.  Then solve the parent problem using the
skills and techniques demonstrated by those examples.  Return a complete
solution and final answer. 
\end{tcolorbox}

\noindent\textbf{Harder leaf synthesis.}\par
\begin{tcolorbox}[apprompt]
\small
\emph{Input:} \texttt{[LEAF PROBLEM]}, \texttt{[VERIFIED SOLUTION]}, and,
when applicable, \texttt{[DOCUMENT]}.

Generate diverse variants that test the same core skills but are harder than
the leaf and require more reasoning steps.  Each variant must be
self-contained and have exactly one verifiable final answer.  Supply a correct
worked solution.  If a document is supplied, use no facts outside it.
\end{tcolorbox}

\noindent\textbf{Bottom-up parent composition.}\par
\begin{tcolorbox}[apprompt]
\small
\emph{Input:} \texttt{[ORIGINAL PARENT--CHILD CONFIGURATION]},
\texttt{[SYNTHETIC CHILD PROBLEMS, SOLUTIONS, AND ANSWERS]}, and, when
applicable, \texttt{[DOCUMENT]}.

Combine the synthetic children into one new parent problem.  Require the
skills from all children and mirror the logical dependency pattern by which
the original children form the original parent.  Match the original parent's
difficulty and level of abstraction; do not simplify.  Return one problem, a
worked solution that composes the child results, and one final answer.  A
document-grounded parent must be answerable solely from the supplied document.
\end{tcolorbox}

\noindent\textbf{Finance document re-grounding.}\par
\begin{tcolorbox}[apprompt]
\small
\emph{Input:} \texttt{[SEED SUMMARY OR SNIPPET]}, \texttt{[SEED ENTITY]},
and \texttt{[NEW DOCUMENT]}.

Locate an analogous entity in the new document and copy a sufficiently
complete supporting passage, preserving its numbers, tables, definitions, and
surrounding conditions.  Re-ground the synthetic tree on this new passage and
entity.  Preserve the reasoning roles of the seed tree, but import no
seed-document facts.  Every resulting node must be answerable from the copied
passage alone.
\end{tcolorbox}

\noindent\textbf{Direct seed synthesis.}\par
\begin{tcolorbox}[apprompt]
\small
\emph{Input:} \texttt{[SEED PROBLEM]}, \texttt{[VERIFIED REASONING TRACE]},
and \texttt{[VERIFIED SOLUTION AND FINAL ANSWER]}.

Generate diverse, self-contained problems that test the same underlying skills identified in the original problem and its verified solution. The problem must be novel and not a simple rewrite. For each problem,
provide a correct worked solution and one final answer.
\end{tcolorbox}

\endgroup

\section{Judge-Based Analysis Protocols}
\label{app:judge-analysis-protocols}

\subsection{Blueprint Preservation and Problem Novelty}

The reported study uses same-seed candidate pairs: one bottom-up
MagicBeanStalk candidate and one direct seed-synthesis candidate.  The two
candidates are selected in the observed Goldilocks band and matched on the
recorded target-policy correct-rollout count.  The reported samples contain
$300$ math pairs and $100$ finance pairs.  A fixed reasoning blueprint is first
extracted from the seed problem, its verified solution, and its reasoning trace.  Candidate scoring by GPT-5.6 is then symmetric and
method-label blinded. The judge receives the seed, the fixed blueprint, and
one generated problem with its reference solution and answer, but no method
name.  The two methods are scored in separate calls under the same rubric.  The
synthetic MBS tree is not exposed during candidate scoring.

\paragraph{Reasoning preservation (1--5).}
\begin{enumerate}[leftmargin=*,nosep]
  \item The required reasoning is different; any overlap is only topical or
  surface-level.
  \item One minor or peripheral component overlaps, but the main reasoning is
  not preserved.
  \item A meaningful subset of the core operations is required, with an
  important operation or dependency omitted or changed.
  \item Most core operations and their principal dependencies are required;
  additional reasoning is allowed.
  \item The same core operations and their composition are clearly required by
  the generated problem.
\end{enumerate}

\paragraph{Problem novelty (1--5).}
\begin{enumerate}[leftmargin=*,nosep]
  \item The candidate is effectively a duplicate or paraphrase of the seed.
  \item The candidate mainly substitutes numbers, names, entities, documents,
  line items, or superficial context.
  \item The candidate is a meaningful variation of substantially the same
  setup.
  \item The candidate is a substantively new setup that still instantiates the
  seed reasoning.
  \item The candidate is a clearly new construction or composition that cannot
  be recovered through superficial substitutions.
\end{enumerate}

The scores are assigned independently. Novelty without blueprint preservation
does not count as successful transfer.  The joint criterion reported in the
paper is met only when both reasoning preservation and problem novelty are at
least $4$ for the same candidate.

\section{MBS Additional Ablation}
\textbf{Which MBS components matter?}
Table~\ref{tab:mbs_component_ablation} evaluates four design choices at their
corresponding construction stage.  With the same four teacher calls per seed,
the heterogeneous teacher pool covers more seeds than the strongest single
teacher.  Teacher verification alone retains all 22 valid decompositions,
whereas the effectivity test removes the 16 whose solutions do not improve
parent solving.  Explicitly requesting harder leaf variants increases
rollout-qualified coverage in the controlled construction audit, and recursive
composition adds qualified problems from 116 source seeds not covered by
one-level composition.

\begin{table*}[t]
  \centering
  \caption{Component analyses.  Teacher coverage uses 100 hard seeds under an
  equal four-call budget.  The leaf comparison uses 10 fixed seeds.  Qualified
  problems produce mixed outcomes over $N=8$ \gptoss rollouts and at least one
  complete correct response within 16K.}
  \label{tab:mbs_component_ablation}
  \small
  \setlength{\tabcolsep}{5.5pt}
  \resizebox{\columnwidth}{!}{%
  \begin{tabular}{@{}llll@{}}
    \toprule
    Component & Without component & With component & Change \\
    \midrule
    Teacher pool
      & Single teacher: $63.0\%$
      & Equal-call pool: $71.5\%$
      & $+8.5$ pp seed coverage \\
    Effectivity test
      & 22 teacher-valid decompositions
      & 6 effective decompositions
      & 16 non-helpful decompositions removed \\
    Harder-leaf synthesis
      & Ordinary variants: $1/10$
      & Harder variants: $3/10$
      & $+2$ qualified seeds \\
    Recursive depth
      & $D=1$: 1,345 roots / 840 seeds
      & $D\leq3$: 1,533 roots / 956 seeds
      & $+188$ roots / $+116$ seeds \\
    \bottomrule
  \end{tabular}
  }
\end{table*}

\end{document}